\documentclass[lettersize,journal]{IEEEtran}

\usepackage[T1]{fontenc}
\usepackage{amsmath,amsfonts,amssymb}
\usepackage{array}
\usepackage[caption=false,font=normalsize,labelfont=sf,textfont=sf]{subfig}
\usepackage{textcomp}
\usepackage{stfloats}
\usepackage{url}
\usepackage{graphicx}
\usepackage{cite}
\usepackage{booktabs}
\usepackage{multirow}
\usepackage{xcolor}
\usepackage{placeins}

\begin{document}

\title{Multipath Adaptive Gated Bottleneck Latent ODE with Raman Data Fusion for Cell Culture Process Forecasting}

\author{Johnny Peng, Thanh Tung Khuat, Ellen Otte, Katarzyna Musial, and Bogdan Gabrys%
\thanks{Johnny Peng, Thanh Tung Khuat, Katarzyna Musial, and Bogdan Gabrys are with the Complex Adaptive Systems Laboratory, The Data Science Institute, University of Technology Sydney, NSW 2007, Australia
(e-mail: johnny.peng@student.uts.edu.au; thanhtung.khuat@uts.edu.au; Katarzyna.Musial-Gabrys@uts.edu.au; bogdan.gabrys@uts.edu.au).}%
\thanks{Ellen Otte is with CSL Innovation, Melbourne, VIC 3000, Australia
(e-mail: ellen.otte@csl.com.au).}%
\thanks{Corresponding author: Johnny Peng.}%
}


\markboth{IEEE Transactions on Neural Networks and Learning Systems}{Anonymous Authors: Multi-Path Adaptive Gated Bottleneck Latent ODE}

\maketitle
\thispagestyle{plain}
\pagestyle{plain}

\begin{abstract}
Mammalian cell-culture processes underpin the manufacture of many biopharmaceuticals, yet keeping a run on track is hard: critical process parameters drift over days, and an off-specification trend is often confirmed too late to intervene. Early-stage, multi-day forecasts could enable timely adjustment of feeding, sampling, and control, but bioprocess forecasting is challenging because measurements are sparse and irregularly sampled, operating conditions are heterogeneous across cell lines and media, and runs with near-identical early behaviour can diverge into different futures. We propose an adaptive framework combining a Gated Bottleneck Latent Ordinary Differential Equation (GB-Latent ODE) with Multi-Path Just-In-Time Fine-Tuning (MP-JIT-FT). The GB-Latent ODE augments the standard Latent ODE with learnable variable-wise gating and a mask-aware bottleneck that compresses high-dimensional sparse inputs, improving learning under limited data. Given a partially observed run, MP-JIT-FT retrieves similar historical trajectories, clusters the local neighbourhood into candidate regimes, and fine-tunes a separate model per regime to produce multiple plausible paths, each with a reconstruction-based confidence score, not a single averaged forecast. We further fuse Raman spectroscopy data: a machine-learning soft sensor turns dense Raman spectra into pseudo-observations that enrich the sparse offline measurements for more robust training. On 38 fed-batch 5L bioreactor runs spanning 14 conditions, MP-JIT-FT with Raman fusion achieves the best average rank and outperforms a global Latent ODE baseline on 8 of 9 target variables. Using local-divergence metrics, we show that the multi-path gains are largest when locally similar prefixes diverge, whereas Raman fusion helps most when early dynamics are representative of later behaviour.
\end{abstract}


\section{Introduction}
\IEEEPARstart{M}{ammalian} cell cultures, and Chinese hamster ovary (CHO) cells in particular, are the dominant production hosts for therapeutic proteins and monoclonal antibodies \cite{khba24,chya20}. Their behaviour is governed by coupled metabolic, environmental, and control dynamics that evolve over a one-to-two-week run, and small early deviations in metabolites such as lactate or ammonia can compound into off-specification batches \cite{brunner2018elevated,gagnon2011high}. Because manual intervention --- adjusting feeding, sampling, or control set-points --- must happen before such trends become irreversible, biopharmaceutical manufacturers increasingly require accurate multi-day forecasts of cell-culture trajectories in the early stages of a run, rather than after the fact \cite{paki23,UNDEY2009177,Jiang2017-kl}.

Producing such forecasts is difficult for several reasons. Bioreactor data are irregularly sampled: sparse offline assays, typically obtained only one to three times per day, are mixed with quasi-continuous online sensing of controlled variables. Datasets are small because each run is expensive and slow to generate \cite{PENG2026108749}. And trajectories can differ sharply across cell lines, media, and control strategies, so models that fit one operating regime may transfer poorly to another. Previous bioprocess forecasting studies have reported promising one-day or multi-step predictions in more homogeneous settings, but performance often degrades when models are deployed under previously unseen operating conditions \cite{zini10,yafi24,jiwa23,paki23}. Prior cell-culture forecasting work leverages Neural ODEs \cite{chru19} because they natively support continuous-time prediction and irregular sampling \cite{Chiu2025}, yet it leaves three practical gaps: it focuses on overly simplified batch settings, assumes limited cross-run heterogeneity, and does not fully exploit mid-run reconditioning from newly observed data.

\begin{figure}[!t]
\centering
\includegraphics[width=1\columnwidth]{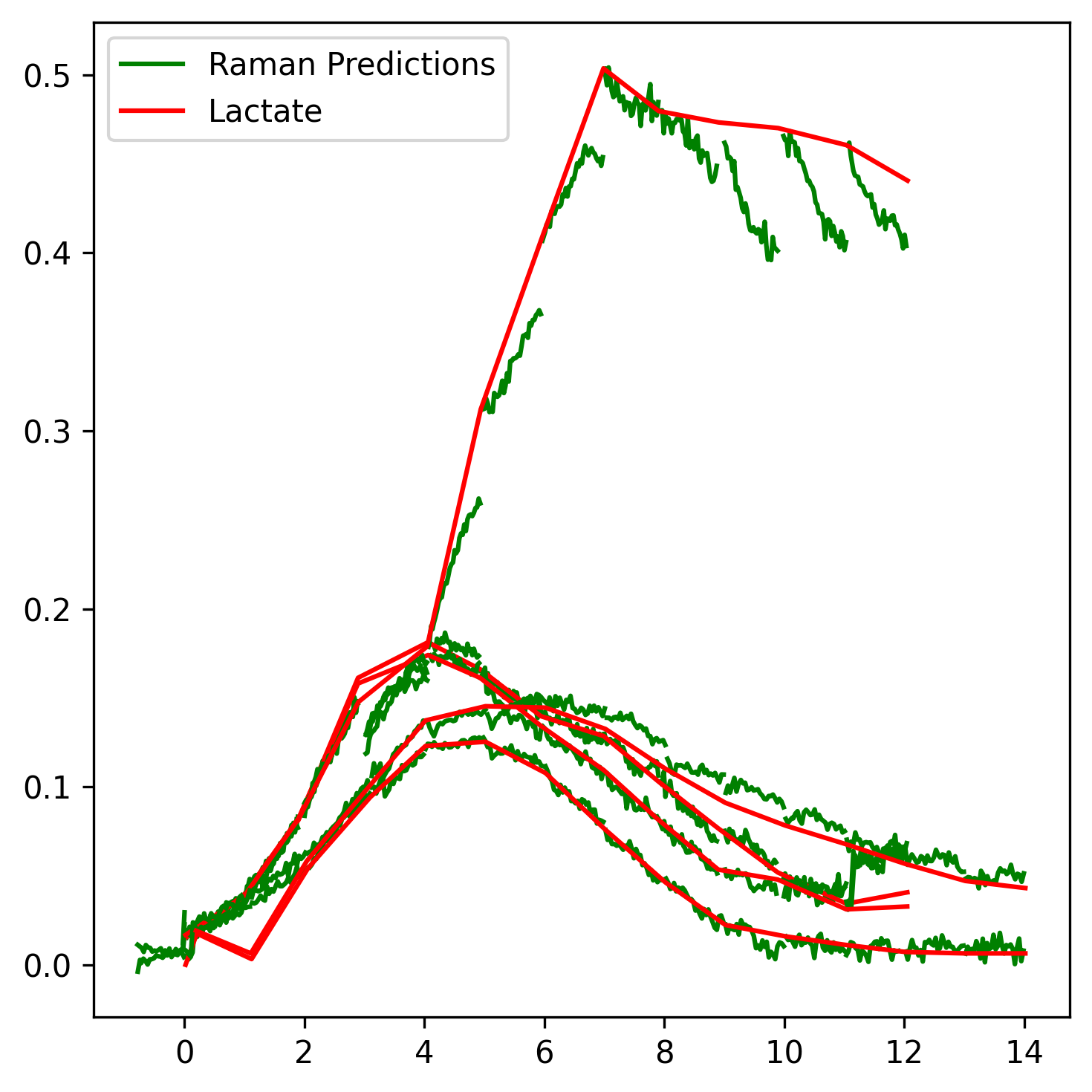}
\caption{Lactate trajectories with similar early values but different future behaviour.}
\label{fig:lactate_traj_short}
\vspace{-10pt}
\end{figure}

This study addresses these limitations using fed-batch data with significant cross-run heterogeneity, which produces runs with similar early prefixes but diverging future outcomes. Fig.~\ref{fig:lactate_traj_short} illustrates this behaviour for lactate: several runs are nearly indistinguishable for the first few days and then separate into distinct regimes. In such cases, a single global model tends to regress toward an ``average'' future that is not operationally useful, and a forecaster that commits to one trajectory cannot express the genuine ambiguity in the data. We argue that this is fundamentally a problem of \emph{multiple-future forecasting under partial observability}, where one observed history may admit several plausible continuations.

To address this, we make three main contributions:
\begin{itemize}
    \item \textbf{Multi-Path Just-In-Time Fine-Tuning (MP-JIT-FT)}, a model-agnostic adaptive framework that retrieves locally similar historical runs, clusters the retrieved neighbourhood into candidate future regimes, and fine-tunes a separate model copy per regime, generating multiple plausible forecasts each scored by a reconstruction-based confidence rather than a single averaged path.
    \item \textbf{Gated Bottleneck Latent ODE (GB-Latent ODE)}, a forecasting model tailored to MP-JIT-FT that adds learnable variable-wise gating and a mask-aware bottleneck on top of the standard Latent ODE \cite{ruch19}, compressing high-dimensional sparse inputs and improving learning when limited data is available \cite{PENG2026108749}. We adopt a Latent ODE backbone because, unlike standard Neural ODEs, it can be adapted to newly observed data mid-run, making it well-suited to just-in-time adaptation.
    \item \textbf{Raman spectroscopy data fusion}, in which a dedicated machine-learning soft sensor converts dense Raman spectra into pseudo-observations of the forecasted variables. This increases the observability of the run --- both before the first offline measurements arrive and between their infrequent recordings --- and we analyse how these estimates can augment sparse offline assays for more reliable training and mid-run adaptation of the GB-Latent ODE.
\end{itemize}

We evaluate the framework on 38 fed-batch 5L bioreactor runs spanning 14 experimental conditions and analyse, using local-divergence metrics, when each component helps. The remainder of the paper reviews the relevant background, presents the method and experimental setup, and reports results and limitations.

\section{Background}
\label{sec:preliminaries}
\subsection{Cell Culture Forecasting}

Modern biopharmaceuticals, and monoclonal antibodies (mAbs) in particular, are predominantly produced in mammalian cell cultures such as CHO cells, and their manufacturing is a complex, tightly regulated, multi-stage process operated under Quality-by-Design (QbD) principles \cite{UNDEY2009177,chya20}. Across the mAb production cycle --- from antibody discovery and clone selection, through upstream process development, to manufacturing and formulation --- machine learning (ML) is increasingly used to extract insight from the large and heterogeneous datasets generated at each stage \cite{khba24}. The upstream cell-culture stage is especially challenging for data-driven modelling, because experiments are costly and slow to run, so the available datasets are typically small and high-dimensional \cite{PENG2026108749}.

A central enabler of data-driven monitoring and control in this setting is Process Analytical Technology (PAT), a regulatory and engineering framework that promotes real-time measurement of critical process parameters and quality attributes to support timely decision-making \cite{UNDEY2009177,Jiang2017-kl}. Within the PAT toolbox, soft sensors --- data-driven inferential models that estimate hard-to-measure variables from readily available online measurements --- have become widely adopted because they reduce reliance on slow, expensive, or destructive offline assays \cite{kagr11,kaga09a}. In cell-culture processes specifically, spectroscopic soft sensors based on Raman spectroscopy are now routinely used to estimate metabolite and biomass concentrations in real time \cite{Tulsyan2019,Tulsyan2020406,Rashedi2024-zd,Poth2022}, and adaptive and just-in-time-learning soft sensors have recently been proposed for the small-data, non-stationary conditions of cell-culture monitoring \cite{PENG2025157,peng2026learning,khba24a,khba24b,khpe25}.

It is important to distinguish soft sensing from forecasting. A soft sensor is, by construction, a \emph{current-time} estimator: it infers the present value of a variable that cannot be measured directly, or is expensive to measure, from other variables observed at the same time \cite{kagr11,baga17}. As such, established soft sensors describe the current process state but do not, in general, predict how that state will evolve. Proactive intervention --- adjusting feeding, sampling, or control set-points before a run degrades --- instead requires anticipating the process trajectory several days ahead, that is, multistep-ahead forecasting, which conventional soft-sensing formulations do not typically cover \cite{paki23}. This gap motivates the multi-day forecasting problem studied in this work.

In fed-batch bioprocessing, key process variables such as lactate and ammonia are typically measured offline only 1--2 times per day, resulting in sparse and irregular observations.

Given a run \(r\) with irregular observation times
\[
\mathcal{T}_{r}=\{t_{r,1}<\cdots<t_{r,n_r}\},
\]
and scalar target \(y_r(t)\), we consider forecasting from an arbitrary continuous
forecast origin \(c\in\mathbb{R}_{\ge 0}\). The observed prefix is
\[
\{(y_r(t_{r,j}),t_{r,j}):t_{r,j}\le c\},
\]
Let
\(\mathcal{H}=\{\tau_1,\ldots,\tau_H\}\) denote the forecast horizons. We adopt a
multi-input multi-output (MIMO) forecaster
\[
\widehat{\mathbf y}_r(c+\mathcal{H})
=
f_{\theta}\!\left(
\{(y_r(t_{r,j}),t_{r,j}):t_{r,j}\le c\}
\right)
\in\mathbb{R}^{H},
\]
where
\[
[\widehat{\mathbf y}_r(c+\mathcal{H})]_h
=
\widehat y_r(c+\tau_h).
\]
The model parameters are learned by minimising the aggregate forecasting loss
across runs and available future target observations:
\[
\min_{\theta}
\sum_{r=1}^{R}
\sum_{\tau\in\mathcal{H}_{r}(c)}
\ell\!\left(
\widehat y_r(c+\tau),
y_r(c+\tau)
\right),
\]
where \(\mathcal{H}_{r}(c)\) denotes the set of evaluated future horizons for which
ground-truth target measurements are available, and \(R\) denotes the total number of bioprocess runs used for model training or evaluation. We use the MIMO strategy because it has previously shown strong performance for multi-step forecasting in cell culture applications \cite{paki23}.

\subsection{Just-in-Time Learning}
Just-in-time learning (JITL) constructs a query-specific local model rather than relying solely on a single global model. For a target query, relevant historical samples are retrieved based on feature-space similarity. In this work, similarity is computed from Euclidean distance,
\[
d(\mathbf x_i,\mathbf x_j)=\|\mathbf x_i-\mathbf x_j\|_2,
\qquad
s(\mathbf x_i,\mathbf x_j)=\exp\!\big(-d(\mathbf x_i,\mathbf x_j)\big),
\]
where $\mathbf x_i,\mathbf x_j\in\mathbb R^n$ are feature vectors. This localised training is intended to improve adaptation under inter-run heterogeneity.

\subsection{Neural ODE and Latent ODE}
Standard Recurrent Neural Networks (RNN) update hidden states only at discrete observation times, which is limiting for irregularly sampled data. Neural ODEs, instead, model hidden-state evolution in continuous time:
\[
\frac{d\mathbf h(t)}{dt}=f_\theta(\mathbf h(t), t),
\qquad
\mathbf h(t_1)=\mathbf h(t_0)+\int_{t_0}^{t_1} f_\theta(\mathbf h(t), t)\,dt.
\]
This formulation naturally supports irregular timestamps by allowing the latent state to be evaluated at arbitrary times. Latent ODE extends this idea to partially observed time series by learning a latent trajectory $\mathbf z(t)$ together with an encoder that infers the latent state from observations, and a decoder that maps latent states back to the observation space \cite{chru19}. To better handle irregular measurements, the encoder can be implemented as an ODE-RNN, which evolves the hidden state continuously between observations and applies a recurrent update when a new observation arrives:
\begin{align}
h^-_i &= \mathrm{ODESolve}(f_\theta, h_{i-1}, (t_{i-1}, t_i)),\\
h_i &= \mathrm{RNNCell}(h^-_i, x_i).
\end{align}
This hybrid design preserves both observation values and time gaps, making it well-suited to sparse and irregular bioprocess trajectories. Our method builds on this latent modelling framework.

\section{Method}

\subsection{Forecasting Setting}

For a run \(r\), different data streams may be observed at different rates. Let
\(\mathcal{T}^{y}_{r}\) denote the sparse offline assay times, \(\mathcal{T}^{u}_{r}\)
the controlled/online-variable times, and \(\mathcal{T}^{\rho}_{r}\) the Raman
pseudo-observation times (only when data fusion is applied). We define the model input grid as the sorted union
\[
\mathcal{T}^{x}_{r}
=
\mathcal{T}^{y}_{r}
\cup
\mathcal{T}^{u}_{r}
\cup
\mathcal{T}^{\rho}_{r}
=
\{s_{r,1}<\cdots<s_{r,n_x}\},
\]
with \(\mathcal{T}^{\rho}_{r}=\emptyset\) when Raman pseudo-observation was not used. At each \(s_{r,i}\in\mathcal{T}^{x}_{r}\), we form
\[
x_r(s_{r,i})
=
[\bar y_r(s_{r,i}),\bar u_r(s_{r,i}),\bar\rho_r(s_{r,i})],
\]
where \(\bar y_r\in\mathbb{R}^{D_y}\) contains target variables,
\(\bar u_r\in\mathbb{R}^{D_u}\) contains controlled or online variables, and
\(\bar\rho_r\in\mathbb{R}^{D_\rho}\) contains Raman pseudo-observations when
available. Entries not observed at \(s_{r,i}\) are treated as missing and represented
by a binary mask \(m_r(s_{r,i})\in\{0,1\}^{D_x}\). This union-time representation
allows sparse offline assays, denser process variables, and Raman pseudo-observations
to enter the model at their native sampling rates.

At a continuous forecast origin \(c\), the model receives the observed prefix
\[
X_r(c)
=
\{(x_r(t),m_r(t),t):t\in\mathcal{T}^{x}_{r},\,t\le c\},
\]
and predicts future target values at horizons
\(\mathcal{H}=\{\tau_1,\ldots,\tau_H\}\):
\[
\widehat{Y}_r(c+\mathcal{H})
=
f_{\theta}(X_r(c)),
\qquad
[\widehat{Y}_r(c+\mathcal{H})]_h
=
\widehat y_r(c+\tau_h)\in\mathbb{R}^{D_y}.
\]
Thus, each target is forecast from the joint history of all available target,
controlled, online, and optional Raman-derived inputs observed before the forecast
origin. In this study, we evaluate forecasts from Day 3 of each run by setting the
continuous forecast origin to \(c=3\) days after inoculation. This timing is chosen because CQAs such as lactate can diverge sharply after Day 4, as shown in Fig.~\ref{fig:lactate_traj_short}, which may already be too late for manual intervention to rescue the run. The prefix, therefore,
contains all observations with timestamps \(t\le c\). Since each run typically lasts for around two weeks, the maximum prediction horizon $\tau_h$ is usually approximately 11 days.

\subsection{Gated Bottleneck Latent ODE}

\begin{figure*}[!t]
\centering
\includegraphics[width=1\textwidth]{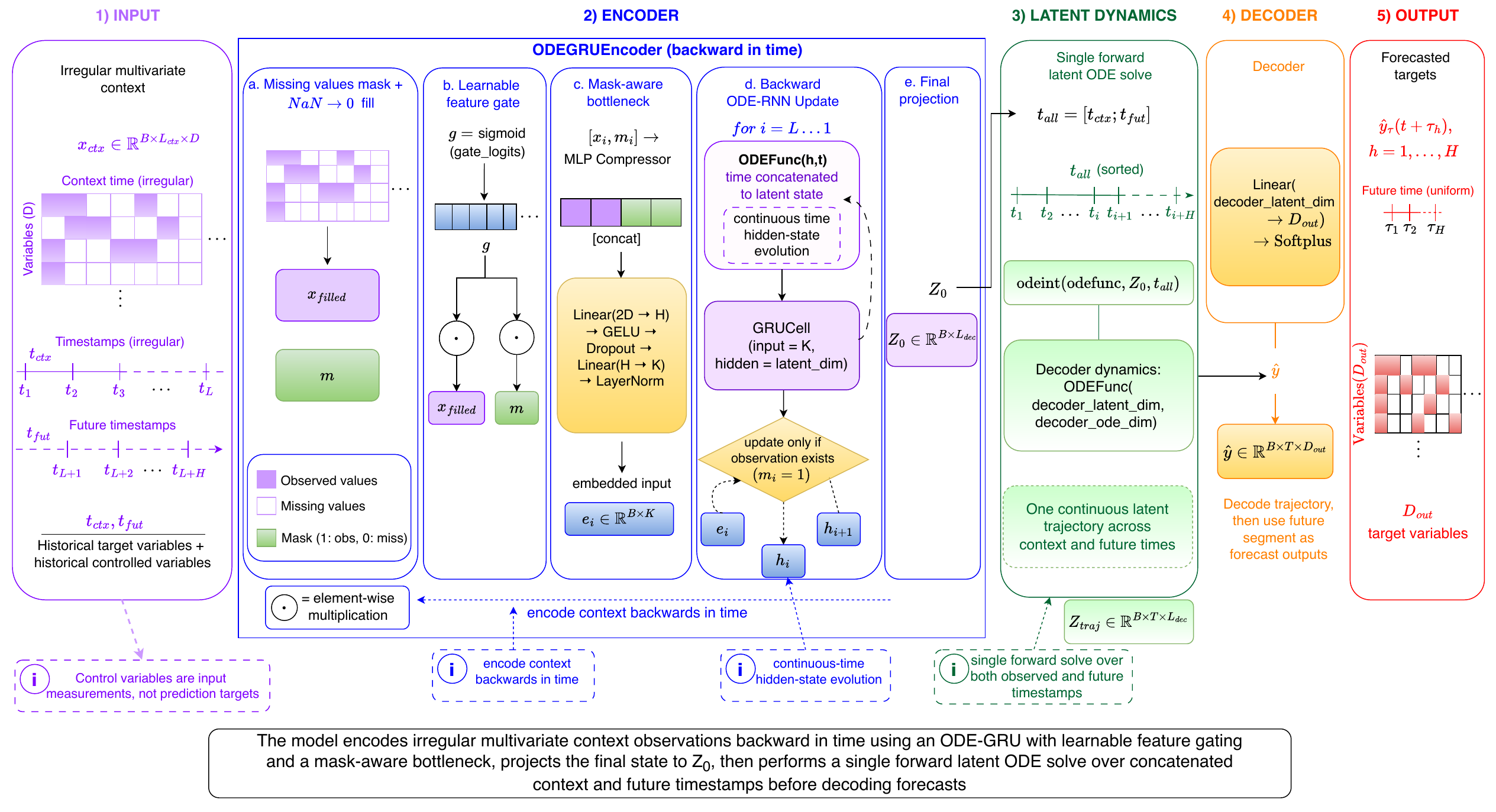}
\caption{Gated Bottleneck Latent ODE.}
\label{fig:model-architecture}
\vspace{-10pt}
\end{figure*}

Latent ODE (with ODE-RNN encoder) combines continuous-time latent dynamics with discrete observation updates, making it well-suited to irregularly sampled time series as illustrated in \cite{ruch19}. Building on this foundation, we propose and develop a \textbf{Gated Bottleneck Latent ODE (GB-Latent ODE)} tailored to multivariate bioprocess forecasting with high-dimensional features and limited data. More specifically, GB-Latent ODE combines the following architectural choices and task-specific adaptations:

\begin{enumerate}

    \item \textbf{Static variable-wise soft gating.}
    Prior to recurrent encoding, each input variable is modulated by a learnable gate
    \(g=\sigma(a)\in\mathbb{R}^{D}\), shared across batch and time. Motivated by
    embedded feature and variable selection methods
    \cite{yamada2020feature,lim2021temporal}, this gate softly reweights input
    channels and is applied to both filled observation values and missingness masks
    before mask-aware bottleneck compression. Thus, it should be interpreted as a
    static channel-calibration mechanism before compression, not as a time-varying
    stage-dependent gating or hard sparsity. The bottleneck embedding discussed below is instead responsible for learning a compact representation of
    the gated value-mask inputs.

    \item \textbf{Compressed bottleneck embedding before recurrent updates.}
    Instead of feeding the raw concatenated value-mask vector directly into the GRU update, we first project it into a lower-dimensional learned embedding. This bottleneck reduces the dimensionality of sparse multivariate inputs into a compact and informative feature representation, which may improve model generalisation when learning from small datasets \cite{PENG2026108749}.

    \item \textbf{Absolute-time conditioning of the ODE dynamics.}
    We implement the latent dynamics in a non-autonomous form by concatenating absolute process time to the latent state at each ODE evaluation in both the encoder and decoder. This is particularly suitable here because cell-culture dynamics are strongly phase- and time-dependent.

    \item \textbf{Activation and output constraints.}
    We replace all Tanh activation functions with GELU \cite{hegi16}, which gives the model greater expressive power to model the sudden changes during fed-batch bioreactor runs resulting from feed additions. Furthermore, we used a Softplus output head to enforce non-negativity for biological targets.

    \item \textbf{Backward encoding with full-trajectory forward decoding.}
    Following the Latent ODE formulation, we encode the observed context backward in time to infer the latent initial state. We then solve a forward latent trajectory over both the observed prefix and future horizon, enabling joint reconstruction and extrapolation. The reconstruction error on the observed prefix is then served as a practical reliability signal for each forecast path in MP-JIT-FT.

    \item \textbf{Deterministic and full-trajectory training objective.}
    Instead of variational latent inference and ELBO optimisation \cite{ruch19}, we adopt a deterministic latent representation and train directly with reconstruction
    and forecasting mean-squared-error losses. Let \(c\) denote the forecast origin and
    let \(\{t_i\}_{i=1}^{N}\) denote the supervised timestamps used in a training
    trajectory. The total loss is
\[
\begin{aligned}
\mathcal{L}_{\mathrm{total}}
&=
\sum_{t_i\le c}
\lVert y_i-\widehat y_i \rVert^{2}
+
\sum_{t_i>c}
\lVert y_i-\widehat y_i \rVert^{2} \\
&=
\mathcal{L}_{\mathrm{recon}}
+
\mathcal{L}_{\mathrm{forecast}} .
\end{aligned}
\]
    This objective is adopted because probabilistic prediction is less critical when MP-JIT-FT models already generate multi-path forecasts. While directly optimising squared error is better aligned with point-forecast accuracy \cite{gneiting2011making}, it avoids the ELBO trade-off between reconstruction fidelity and prior regularisation \cite{lin2019balancing}, and reduces exposure to VAE failure modes such as posterior collapse \cite{he2019lagging}.
\end{enumerate}
While not all components listed above are individually novel, our contribution lies in the task-specific integration of them for the application of cell-culture process forecasting. Fig.~\ref{fig:model-architecture} illustrates the resulting architecture, and we now describe its forward computation in detail, following the exact computational flow used in our implementation.

\subsubsection{Input representation}
Let \(x_{\mathrm{ctx}}\in\mathbb{R}^{B\times L_{\mathrm{ctx}}\times D}\) denote a batch of multivariate context trajectories observed up to the forecast origin. To avoid ambiguity between context and future timestamps, we denote the observed context timestamps by \(c_{1:L_{\mathrm{ctx}}}=(c_1,\ldots,c_{L_{\mathrm{ctx}}})\) and the future query timestamps by \(q_{1:L_{\mathrm{fut}}}=(q_1,\ldots,q_{L_{\mathrm{fut}}})\), with \(c_1<\cdots<c_{L_{\mathrm{ctx}}}<q_1<\cdots<q_{L_{\mathrm{fut}}}\). The model receives \((x_{\mathrm{ctx}}, c_{1:L_{\mathrm{ctx}}}, q_{1:L_{\mathrm{fut}}})\), but the encoder processes only the observed context timestamps \(c_{1:L_{\mathrm{ctx}}}\) in reverse temporal order; the future query timestamps are used only by the decoder-side forward solve.

\subsubsection{Continuous-time dynamics block}
Both the encoder-side and decoder-side latent dynamics share the same functional form, implemented as an MLP that explicitly concatenates time to the latent state. For a latent state \(\mathbf{h}(t)\in\mathbb{R}^{H}\),
\[
\frac{d\mathbf{h}(t)}{dt}=f_\theta(\mathbf{h}(t),t),
\qquad
f_\theta(\mathbf{h},t)=\mathrm{MLP}\!\left([\mathbf{h};t]\right),
\]
where the MLP is
\[
\begin{aligned}
\relax[\mathbf{h}; t]
&\rightarrow \mathrm{Linear}(H+1, H_{\mathrm{ode}})
\rightarrow \mathrm{GELU} \\
&\rightarrow \mathrm{Linear}(H_{\mathrm{ode}}, H_{\mathrm{ode}})
\rightarrow \mathrm{GELU}
\rightarrow \mathrm{Linear}(H_{\mathrm{ode}}, H_{\mathrm{ode}}) \\
&\rightarrow \mathrm{GELU}
\rightarrow \mathrm{Linear}(H_{\mathrm{ode}}, H).
\end{aligned}
\]
The explicit concatenation of time makes the dynamics non-autonomous, which matters for bioprocess trajectories whose latent evolution depends not only on state but also on the absolute process stage.

\subsubsection{Missing-value handling and feature gating}
The first encoder-side operation constructs a mask indicating which entries are observed, \(\mathbf{m}=\mathbf{1}_{\mathrm{finite}}(\mathbf{x}_{\mathrm{ctx}})\), taking value \(1\) for observed entries and \(0\) for missing ones. Missing values are then replaced by zero, \(\tilde{\mathbf{x}}=\mathrm{nan\_to\_num}(\mathbf{x}_{\mathrm{ctx}},0)\). A learnable variable-wise gate \(\mathbf{g}=\sigma(\mathbf{a})\in\mathbb{R}^{D}\), where \(\mathbf{a}\) is a learned parameter vector and \(\sigma(\cdot)\) is the sigmoid, is then broadcast across batch and time and applied to both the filled values and the mask:
\[
\tilde{\mathbf{x}}^{(g)}=\tilde{\mathbf{x}}\odot\mathbf{g},
\qquad
\mathbf{m}^{(g)}=\mathbf{m}\odot\mathbf{g}.
\]
Applying the same gate to values and masks lets the model learn which channels should contribute, both through their magnitude and through their observed/missing pattern, before the recurrent encoder update.

\subsubsection{Mask-aware bottleneck embedding}
At each context timestamp \(c_i\), the gated values and masks \(\mathbf{x}_i=\tilde{\mathbf{x}}^{(g)}[:,i,:]\) and \(\mathbf{m}_i=\mathbf{m}^{(g)}[:,i,:]\) are concatenated into \(\mathbf{u}_i=[\mathbf{x}_i;\mathbf{m}_i]\in\mathbb{R}^{2D}\). Rather than feeding this \(2D\)-dimensional vector directly into the GRU, the model compresses it through a bottleneck MLP \(\mathbf{e}_i=\phi_{\mathrm{embed}}(\mathbf{u}_i)\in\mathbb{R}^{K}\), where
\[
\begin{aligned}
\phi_{\mathrm{embed}}
&=
\mathrm{LayerNorm}
\circ
\mathrm{Linear}(H_{\mathrm{emb}}, K) \\
&\quad\circ
\mathrm{Dropout}
\circ
\mathrm{GELU}
\circ
\mathrm{Linear}(2D, H_{\mathrm{emb}}),
\end{aligned}
\]
and \(K\) is the bottleneck embedding dimension. This stage explicitly incorporates missingness information into the recurrent input while compressing the high-dimensional concatenated feature-mask vector into a compact representation.

\subsubsection{Backward ODE-GRU encoder}
The encoder is an ODE-RNN module that processes only the observed context in reverse time order, initialised at the last observed context timestamp with \(h^{-}_{L_{\mathrm{ctx}}}=0\). The observation indicator is computed from the original binary mask \(\bar{m}_{i,j}\):
\[
\mathrm{has\_obs}_i
=
\mathbf{1}\!\left(\sum_{j=1}^{D}\bar{m}_{i,j}>0\right).
\]
We use the ungated mask here because \(g_j=\sigma(a_j)>0\), so checking the gated mask would be mathematically equivalent but less transparent. The final context observation is assimilated by
\[
\begin{aligned}
h_{L_{\mathrm{ctx}}}
&=
\mathrm{has\_obs}_{L_{\mathrm{ctx}}}\cdot
\mathrm{GRUCell}(e_{L_{\mathrm{ctx}}}, h^{-}_{L_{\mathrm{ctx}}}) \\
&\quad+(1-\mathrm{has\_obs}_{L_{\mathrm{ctx}}})\cdot h^{-}_{L_{\mathrm{ctx}}}.
\end{aligned}
\]
Then, for \(i=L_{\mathrm{ctx}}-1,\ldots,1\), the hidden state is evolved backward between adjacent observed timestamps and updated on assimilation:
\[
h^{-}_{i}=\mathrm{ODESolve}(f_\theta, h_{i+1}, [c_{i+1},c_i]),
\quad
\]
\[
h^{\mathrm{new}}_i=\mathrm{GRUCell}(e_i,h^{-}_i),
\]
\[
h_i=\mathrm{has\_obs}_i\cdot h^{\mathrm{new}}_i+(1-\mathrm{has\_obs}_i)\cdot h^{-}_i.
\]
If a time point contains no valid observations, the GRU update is skipped, and the ODE-evolved hidden state is kept. The encoder, therefore, depends only on observed context timestamps and is invariant to the choice of future query timestamps.

\subsubsection{Projection, forward solve, and decoding}
After the reverse-time encoder processes the full context, the earliest-time hidden state is mapped to the decoder latent initial condition \(\mathbf{z}_0=\mathbf{W}_z\mathbf{h}_1+\mathbf{b}_z\), a compressed representation of the entire observed context expressed at the earliest context time. The model then constructs a single ascending time vector \(t_{\mathrm{all}}=[c_{1:L_{\mathrm{ctx}}};q_{1:L_{\mathrm{fut}}}]\) and performs \emph{one} continuous solve across both observed and future times:
\[
\frac{d\mathbf{z}(t)}{dt}=g_\psi(\mathbf{z}(t),t),
\quad
\mathbf{z}(t_1)=\mathbf{z}_0,
\quad
\]
\[
\mathbf{Z}_{\mathrm{traj}}=\mathrm{ODESolve}(g_\psi,\mathbf{z}_0,t_{\mathrm{all}}).
\]
A single solve over the entire span keeps the latent trajectory temporally consistent. Each latent state is decoded by \(\mathrm{Decoder}=\mathrm{Softplus}\circ\mathrm{Linear}(H_{\mathrm{dec}},9)\), fixing the output to the \(9\) target variables and enforcing non-negativity. The decoder is applied to the full trajectory, \(\hat{\mathbf{Y}}=\mathrm{Decoder}(\mathbf{Z}_{\mathrm{traj}})\), including the observed-context portion, which is later reused for confidence estimation in MP-JIT-FT.

\subsubsection{Summary of the forward pass}
Putting the components together, the forward pass is
\begin{align}
\mathbf{m} &= \mathbf{1}_{\mathrm{finite}}(\mathbf{x}_{\mathrm{ctx}}), \quad
\tilde{\mathbf{x}} = \mathrm{nan\_to\_num}(\mathbf{x}_{\mathrm{ctx}}, 0), \nonumber \\
\mathbf{g} &= \sigma(\mathbf{a}), \quad
\tilde{\mathbf{x}}^{(g)} = \tilde{\mathbf{x}} \odot \mathbf{g}, \quad
\mathbf{m}^{(g)} = \mathbf{m} \odot \mathbf{g}, \nonumber \\
\mathbf{u}_i &= [\tilde{\mathbf{x}}^{(g)}_i ; \mathbf{m}^{(g)}_i], \quad
\mathbf{e}_i = \phi_{\mathrm{embed}}(\mathbf{u}_i), \nonumber \\
\mathbf{h}_i^{-} &= \mathrm{ODESolve}(f_\theta, \mathbf{h}_{i+1}, [c_{i+1}, c_i]), \nonumber \\
\mathbf{h}_i &= \mathrm{has\_obs}_i \cdot \mathrm{GRUCell}(\mathbf{e}_i,\mathbf{h}_i^{-})
+ (1-\mathrm{has\_obs}_i)\cdot \mathbf{h}_i^{-}, \nonumber \\
\mathbf{z}_0 &= \mathrm{Linear}(\mathbf{h}_1), \quad
t_{\mathrm{all}} = [c_{1:L_{\mathrm{ctx}}};q_{1:L_{\mathrm{fut}}}], \nonumber \\
\mathbf{Z}_{\mathrm{traj}} &= \mathrm{ODESolve}(g_\psi,\mathbf{z}_0,t_{\mathrm{all}}), \quad
\hat{\mathbf{Y}} = \mathrm{Decoder}(\mathbf{Z}_{\mathrm{traj}}).
\end{align}
The encoder and decoder ODE blocks share the same functional form but operate in different latent spaces (the encoder uses the encoder latent dimension and the decoder uses the decoder latent dimension), and both use the Dormand--Prince \texttt{dopri5} solver from \texttt{torchdiffeq}.

\subsection{Multi-Path Just-In-Time Fine-Tuning}
Global models trained on all runs may perform poorly when historical trajectories are highly heterogeneous. JIT transfer learning provides a natural remedy by adapting a pre-trained global model at inference time using locally relevant historical samples, and related JIT Fine-Tuning (JIT-FT) and transfer-learning strategies have shown promising results in adaptive soft sensing \cite{wuli20,dai2023}. Likewise, clustering-based localisation has been widely studied for heterogeneous forecasting problems, where related series are partitioned into more homogeneous groups so that cluster-specific or localised global models can better capture subgroup-specific dynamics than a single pooled model \cite{10.1007/978-3-031-24378-3_2, GODAHEWA2021107518, https://doi.org/10.1002/for.3263}. However, neither approach alone directly addresses the setting considered in this study, where runs may exhibit highly similar behaviour up to the current observation time yet diverge into multiple plausible future regimes thereafter, as illustrated in Fig.~\ref{fig:lactate_traj_short}. Standard JIT-FT adapts a \emph{single} local predictor from a retrieved neighbourhood, so when that neighbourhood contains samples with similar prefixes but conflicting future continuations, it does not explicitly preserve these competing modes and may instead favour one dominant mode or an averaged local solution. Clustering-based localisation alone is also insufficient in this setting, because it is typically used to construct fixed offline groups that reduce \emph{global} heterogeneity, rather than to resolve \emph{query-specific} future ambiguity within the local neighbourhood of the current partially observed run.

\begin{figure*}[!t]
\centering
\includegraphics[width=1\textwidth]{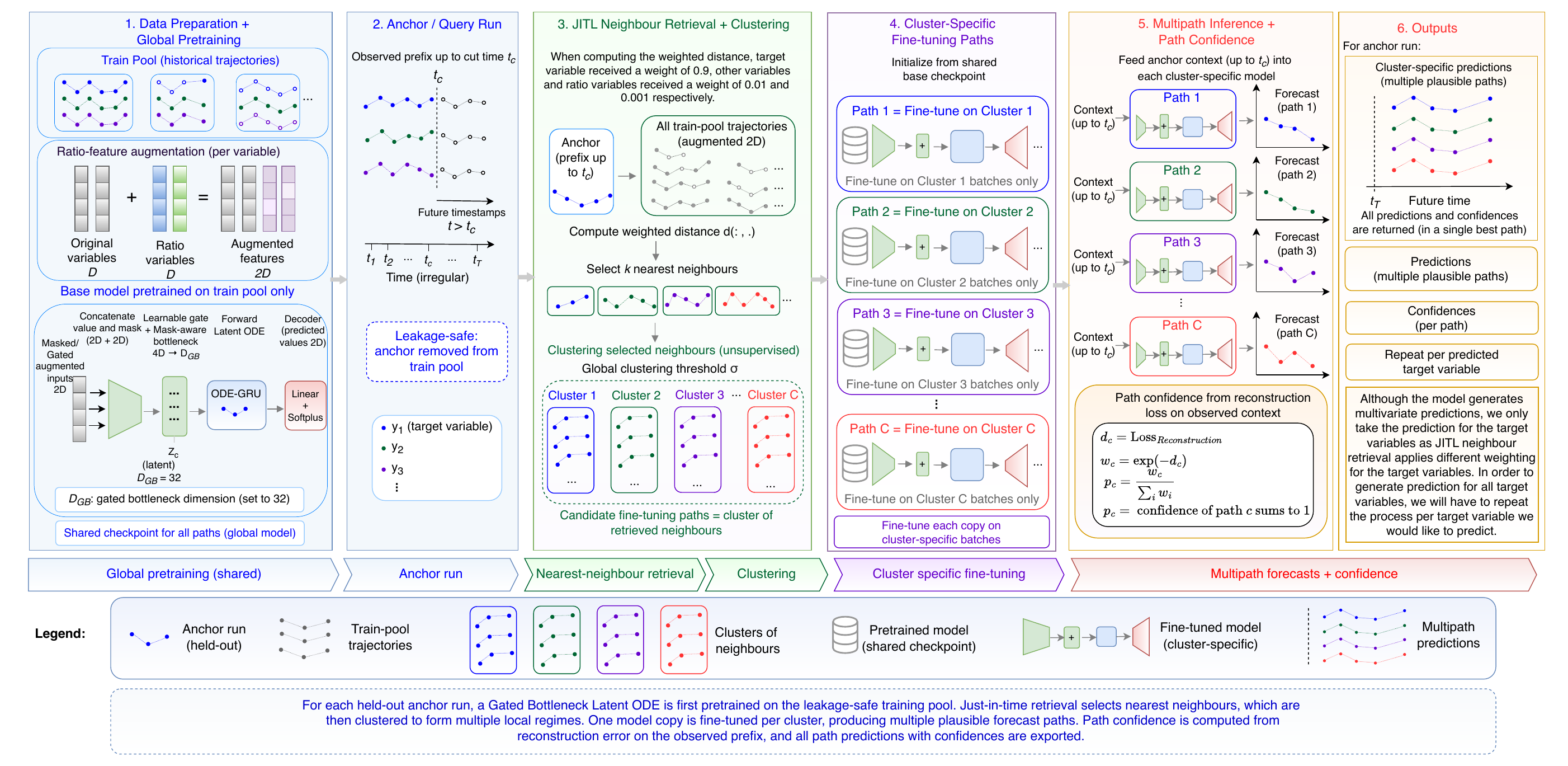}
\caption{Multi-Path Just-In-Time Fine-Tuning pipeline with Gated Bottleneck Latent ODE.}
\label{fig:Pipeline}
\vspace{-10pt}
\end{figure*}

Motivated by these limitations, we propose a \emph{Multi-Path Just-In-Time Fine-Tuning} (MP-JIT-FT) framework, which combines JIT-FT and clustering-based localisation technique by first performing query-specific local retrieval, then clustering the retrieved neighbourhood itself to identify competing future regimes, and finally fine-tuning a pretrained global model separately on each regime to produce multiple path-specific forecasts. More generally, this can be viewed as a new framework for addressing the \emph{multiple-future forecasting under partial observability} problem, where one observed history may admit multiple plausible futures \cite{10.5555/3454287.3455669}. Fig.~\ref{fig:Pipeline} shows a high-level overview of the MP-JIT-FT pipeline, and we describe each stage in detail below.

The first step of the pipeline is to pre-train the global model. For each held-out anchor (query) run \(\mathcal{A}\), the pipeline first removes every trajectory sharing the anchor's experiment identifier from the training pool, \(\mathcal{D}_{\mathrm{train}}=\{\tau\in\mathcal{D}:\mathrm{id}(\tau)\neq\mathrm{id}(\mathcal{A})\}\), so that the anchor influences neither global pretraining nor neighbour retrieval. A global GB-Latent ODE is then pretrained on \(\mathcal{D}_{\mathrm{train}}\) by minimising the mean squared error accumulated over all valid finite target entries,
\[
\mathcal{L}_{\mathrm{pretrain}}
=
\frac{1}{|\mathcal{B}|}\sum_{b\in\mathcal{B}}
\mathrm{MSE}\!\left(\hat{\mathbf{y}}^{(b)},\mathbf{y}^{(b)}_{\mathrm{true}}\right),
\]
yielding a shared checkpoint \(\Theta_{\mathrm{base}}\) that initialises all subsequent fine-tuning paths. Although the model input contains both original and ratio-augmented variables, the supervised target uses only the original process variables.

It is worth noting that the pipeline uses both absolute values and ratio-augmented time-normalised relative-change features as inputs. Absolute values capture the current process state, while ratio features provide rate-like information about recent dynamics. For a trajectory with variable \(x_{i,j}\) observed at irregular timestamps \(t_i\) with \(\Delta t_i = t_i - t_{i-1}\), the relative-change feature is
\[
r_{i,j} = \frac{x_{i,j}-x_{i-1,j}}{|x_{i-1,j}|\,\Delta t_i},
\qquad i\ge 2,
\]
with \(r_{1,j}\), and any entry for which \(x_{i,j}\) or \(x_{i-1,j}\) is missing or \(|x_{i-1,j}|=0\), set to NaN, giving the augmented trajectory \(\tilde{x}_i=[x_i;r_i]\in\mathbb{R}^{2D}\). Unlike a simple sequential ratio \(x_{i,j}/x_{i-1,j}\), dividing by the elapsed time avoids confounding biological change with the irregular sampling schedule, which is more appropriate for sparse bioprocess trajectories. These relative-change channels are used only as auxiliary inputs for representation learning and neighbour retrieval, while the supervised targets remain the original process variables. Using rate-derived quantities is useful for cell-culture modelling because they have been widely used to characterise process state and to predict offline measurements such as viable cell count or biomass-related variables in mammalian and CHO cell-culture processes \cite{winter2024soft,survyla2023viable,bayer2020shortcomings}. Combining both feature types can therefore help distinguish trajectories with similar current values but different recent trends.

Then, we construct the subsets of historical data for fine-tuning based on a partially
observed run. We first linearly interpolate each trajectory prefix onto a common
time grid with $n_{\mathrm{grid}}=100$ equally spaced points. For two trajectories
$a$ and $b$, the common grid is defined over the joint time range of the two
prefixes as $\mathcal{T}_{ab}=\left\{\tau_i\right\}_{i=1}^{n_{\mathrm{grid}}}$. Linear interpolation is then applied independently to each variable over this common grid. In our experiments, the distance is computed over all variables after ratio-feature augmentation, including both the original process variables and their corresponding ratio features. For each prediction target $q$, we use a target-specific weighting scheme, where the target variable is assigned weight $w_m^{(q)}=0.9$, the other original process variables are assigned weight $w_m^{(q)}=0.01$, and ratio features are assigned weight $w_m^{(q)}=0.001$. Thus, neighbour selection is primarily driven by similarity in the target variable, while other process variables and ratio-based trend features provide weaker contextual contributions to the distance.

Let $\tilde{x}_{i,m}^{(a)}$ and $\tilde{x}_{i,m}^{(b)}$ denote the interpolated
values of variable $m$ for trajectories $a$ and $b$ at grid point $\tau_i$.
The target-specific weighted distance used for neighbour selection is computed
as
\[
D_{ab}^{(q)}
=
\sqrt{
\frac{
\sum_{i=1}^{n_{\mathrm{grid}}}
\sum_{m=1}^{M}
w_m^{(q)}
\left(
\tilde{x}_{i,m}^{(a)}
-
\tilde{x}_{i,m}^{(b)}
\right)^2
}{
\sum_{i=1}^{n_{\mathrm{grid}}}
\sum_{m=1}^{M}
w_m^{(q)}
}
},
\]
where $M$ denotes the total number of variables after ratio-feature augmentation. In this study, we retrieve $k=5$ nearest trajectories under this weighted interpolated distance to form the initial reference set for the observed run. In practice, the value of $k$ controls the locality of adaptation and should be chosen according to the underlying data and application.

It is acknowledged that different clustering methods and cutoffs can be used at this step.
In our implementation, we apply hierarchical agglomerative clustering with average linkage to
the distance matrix computed over the retrieved reference set. The resulting dendrogram is
converted into flat clusters using a fixed cophenetic-distance cutoff. Specifically, two candidate neighbours are assigned to the same flat cluster only when their cophenetic distance in the hierarchical tree is no greater than the 90th percentile of all cophenetic distances. This value is treated as a fixed distance-tolerance hyperparameter, held constant across all targets, folds, and model variants to avoid test-set-specific tuning. Its role is to provide a reproducible rule for deciding whether the retrieved neighbours should be fine-tuned as one local regime or separated into multiple candidate future paths. In practice, the percentile threshold should be chosen based on the desired level of locality in the data.


Then, separate forecasting models are fine-tuned on each cluster, producing a set of candidate future paths that reflect different local regimes among the retrieved historical neighbours. Path confidence is computed from the reconstruction loss on the observed prefix as:
\[
W_j =
\frac{\exp(-L^{\text{recon}}_j)}
{\sum_{k=1}^{m}\exp(-L^{\text{recon}}_k)}
\]
where \(j \in \{1,\ldots,m\}\) indexes the candidate paths, \(m\) is the number of candidate paths generated by clustering the retrieved neighbours, and \(N_j\) is the number of valid observed-prefix points used to compute the reconstruction loss for path \(j\). Here, \(\hat{y}_{j,n}\) denotes the reconstruction produced by the \(j\)-th cluster-specific fine-tuned model, while \(y_n\) denotes the corresponding observed value. Note that all variables are range-normalised before model fitting, thus the confidence computation is not affected by the magnitudes of different variables. For a model-agnostic alternative, path confidence can also be obtained by inverse-distance weighting of member runs: the pipeline forms neighbour-level priors \(\pi_i=\exp(-d_i)/\sum_{j=1}^{k}\exp(-d_j)\), where \(d_i\) is the distance of the \(i\)-th neighbour to the anchor, and aggregates them at the cluster level to give each cluster an initial prior mass. However, reconstruction-loss-based confidence is more informative than simple distance-based weighting because it reflects both local data heterogeneity and how well each fine-tuned model explains the observed prefix.

At inference, the anchor trajectory is split into context and future segments at the forecast origin, and each cluster-specific model is evaluated on the same anchor context to produce one forecast trajectory per cluster, \(\{\hat{\mathbf{Y}}^{(1)},\ldots,\hat{\mathbf{Y}}^{(M)}\}\). Rather than committing to a single winner, the pipeline exports all path predictions together with their per-target reconstruction-based confidences. Conceptually, MP-JIT-FT therefore combines three ideas: \emph{global representation learning}, in which the base GB-Latent ODE learns a generic mapping from irregular multivariate prefixes to full trajectories; \emph{query-specific localisation}, in which JITL retrieves the historical trajectories most relevant to the current anchor and target variable; and \emph{multipath adaptation}, in which clustering splits the retrieved neighbours into local regimes, each yielding one fine-tuned model. Unlike a conventional single-path JITL pipeline that collapses all retrieved neighbours into one adaptation set, MP-JIT-FT preserves multiple locally coherent alternatives and ranks them by observed-prefix reconstruction quality.


\subsection{Raman Spectroscopy Data Fusion}
\begin{figure*}[!t]
\centering
\includegraphics[width=1\textwidth]{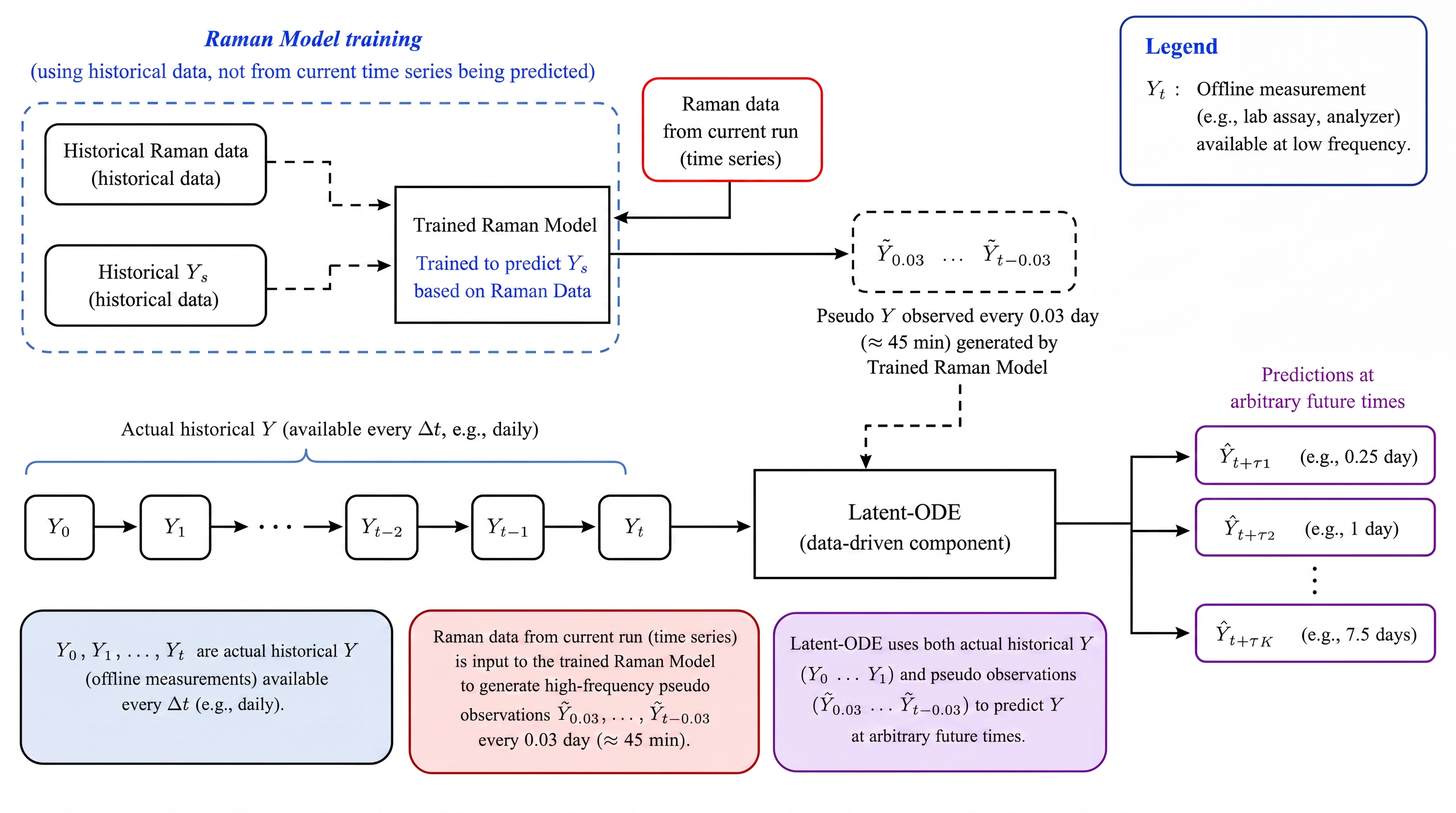}
\caption{Raman Data Fusion via Pseduo-Observation}
\label{fig:data fusion}
\vspace{-10pt}
\end{figure*}

Offline assays are sparse but more accurate, while Raman spectra-based predictions, obtained from the machine learning based soft sensor, are denser but noisier \cite{peng2026learning}. Thus, we can fuse them by treating Raman spectra-based predictions as pseudo-observations during training. As shown in Fig~\ref{fig:data fusion}, let $\{(t_i,\mathbf y_i)\}_{i=1}^{n}$ be the offline measurements and let a pretrained Raman model $g_\phi$ produce pseudo-observations $\tilde{\mathbf y}_j=g_\phi(\mathrm{Raman}(\tau_j))$ on a dense grid $\{\tau_j\}_{j=1}^{m}$. The GB-Latent ODE is trained on the union of true and pseudo-observations with the loss function below:
\[
\mathcal L(\theta)=\frac{1}{n}\sum_{i=1}^{n}\ell\bigl(\hat{\mathbf y}(t_i),\mathbf y_i\bigr)
+\lambda\frac{1}{m}\sum_{j=1}^{m}\ell\bigl(\hat{\mathbf y}(\tau_j),\tilde{\mathbf y}_j\bigr),
\]
Compared with directly adding Raman spectra to the forecasting model, this pseudo-observation data fusion approach avoids introducing more than 3,000 Raman wavenumber features, thereby reducing computational cost and the risk of overfitting under limited data.

In this study, we set \(\lambda=1\) to isolate whether Raman-derived pseudo-observations can provide useful temporal information without introducing an additional tuning procedure. However, this choice should be interpreted as a fixed baseline weighting rather than an optimal treatment of pseudo-observation reliability. Because Raman pseudo-observations are predictions from a separate soft-sensor model, their errors may propagate into the forecasting model and can bias the learned latent dynamics. Future study could extend this by developing a target- or uncertainty-dependent \(\lambda_{j}\).

\section{Experimental Setup}
\subsection{Data and Preprocessing}
The dataset contains 38 fed-batch 5L bioreactor runs spanning 14 conditions that differ in cell line, feed, and fill-media composition. Each run lasts roughly 1 to 2 weeks, and 1 to 3 offline measurements are collected per day for 10 variables, including glucose, lactate, ammonia, viable cell density (VCD), glutamate, glutamine, calcium, sodium, potassium, and osmolality. Nine of the ten offline-measured variables, excluding glucose, serve as target variables for our model. Glucose is excluded because it is added to the bioreactor as part of the daily feeding process, which introduces substantial spikes in glucose levels throughout the run. We do not have time information on when these spikes occur, making glucose unpredictable and fundamentally violating the forecast signal continuity condition required by the Neural ODE models (i.e. the glucose values have multiple sudden changes resulting from the manual interventions which are common in fed-batch processes).

The input features include all past observations of all offline-measured variables (including glucose) and controlled variables, including Temperature, Dissolved oxygen, pH, Mixing Speed, Oxygen, Carbon Dioxide, Sparger Air, Overlay Air, and Head Space Air. Raman pseudo-observations are predictions from a separate Raman spectra-based ML model trained on Raman spectra, available approximately every 45 minutes. The Raman data are preprocessed using the steps outlined in a recent benchmark study \cite{peng2026learning}. The model with the highest rank in that benchmark study was used to generate Raman predictions for each target variable. All input data are normalised using min-max normalisation before model fitting, and the Raman models are learned using the JITL method with $k=100$, which fits the model at inference time using the 100 nearest data points.

\subsection{Benchmark Models and Evaluation}
We compare the proposed MP-JIT-FT model variants against three global baseline models trained on all available historical data: a Linear Kalman Filter (LKF) \cite{kalman}, Classification and Regression Trees (CART) \cite{breiman1984classification}, and the original Latent ODE \cite{ruch19}. Univariate LKF was included as a simple baseline because it naturally handles irregular timestamps and uses only the lagged values of the target variables, representing the minimum level of model performance. CART was included because it achieved the best overall performance in a recent benchmark study on multi-step forecasting for cell-culture process data \cite{paki23}. As CART does not natively handle irregular timestamps, the data were first average-pooled into daily intervals for training, and its daily predictions were then linearly interpolated to irregular time points for evaluation. Latent ODE was included because it forms the basis of our proposed method and, like LKF, can natively handle irregular timestamps while capturing more complex dynamics. For all MP-JIT-FT models, we used the prediction path with the highest confidence for evaluation. The exact hyperparameters used for each model are detailed in Section~\ref{subsec:hyperparams}.

Although there are other more recent models for irregularly sampled multivariate time-series forecasting, we did not include them in the present evaluation because the primary focus of this paper is not to provide an exhaustive benchmark of irregular-time forecasting architectures, but to evaluate whether the proposed model-agnostic MP-JIT-FT framework can improve forecasting when locally similar prefixes lead to divergent future trajectories. We therefore use Latent ODE as the main baseline and leave a broader comparison as an important direction for future work.


Models are evaluated using leave-one-batch-out cross-validation, with each batch comprising 1--4 runs generated under similar conditions. Predictive performance is quantified by range-Normalised Mean Absolute Error (NMAE), while confidence intervals were constructed using the cross-validation variance estimators from \cite{NEURIPS2020_bce9abf2}. In our setting, fold sizes varied across runs, so we adapted their estimators to the observation level rather than treating all folds equally. Specifically, for the all-pairs estimator, we pooled all held-out pointwise losses across folds and estimated the asymptotic variance from their squared deviations around the overall cross-validation mean; for the within-fold estimator, we replaced the equal-fold average with a size-weighted average of the within-fold sample variances, weighting each fold by its number of held-out observations. This modification reduces to the original Bayle estimator \cite{NEURIPS2020_bce9abf2} when all folds have the same size, but for unequal fold sizes, it should be viewed as a practical extension rather than an exact result covered by their theory.

\subsection{Model Hyperparameters and implementations}
\label{subsec:hyperparams}
For all Latent ODE models in this study, including both the original Latent ODE and the proposed GB-Latent ODE, we used one of the hyperparameter sets that were used in the original Latent ODE paper \cite{chru19}: Learning Rate was set to 0.01, Generative/Decoder Model Latent dimensions was set to 20, Recognition/Encoder Model Latent dimensions was set to 40, ODE functions have 3 layers with 50 units each. Training uses Adamax with learning rate lr = 0.01. Original Latent ODE was trained for 2000 iterations, while the MP-JIT-FT models were first globally pretrained for 1000 iterations on all non-held-out trajectories, then fine-tuned for 1000 iterations separately on each JITL cluster. When selecting $k=5$ nearest neighbours for JITL, the target variable receives a weight of 0.9, and all other non-ratio variables receive a weight of 0.01, while all ratio features receive a weight of 0.001. Furthermore, GB-Latent ODE's encoder uses a mask-aware bottleneck compressor that maps the concatenated gated values and missingness masks from dimension 2D (D = Input Feature Dimension Size) to a compact embedding of 32 dimensions, using a one-hidden-layer MLP with hidden size 64, GELU activation, dropout 0.1, and LayerNorm. In the ratio-augmented setting, D=20, so the compressor maps a 40-dimensional input to a 32-dimensional bottleneck embedding before passing it into the ODE-GRU encoder.

For the CART model, following the hyperparameter setting from the original benchmark paper \cite{paki23}, we have used the default model hyperparameters in Python's sklearn implementation of CART. For the LKF, we use a 2\textendash state local\textendash level + local\textendash trend model for univariate observations with fixed transition \(A(\Delta t)=\begin{bmatrix}1&\Delta t\\0&1\end{bmatrix}\) and observation \(H=[1,\,0]\). The process\textendash noise covariance is diagonal, \(Q=\mathrm{diag}(\sigma^2_{q,\text{level}},\,\sigma^2_{q,\text{trend}})\), and the observation\textendash noise covariance is scalar, \(R=\sigma_r^2\); the corresponding noise magnitudes are learned via softplus\textendash parametrized log\textendash standard deviations to ensure positivity. The initial state is set from data, \(m_0=\big[x_0,\ (x_1-x_0)/(t_1-t_0)\big]^\top\) (zero trend if only one point is available), with initial covariance \(P_0=\mathrm{diag}(1,1)\). Training used a learning rate of \(5\times 10^{-3}\) for 140 epochs.

All experiments were implemented using standard open-source scientific Python software. CART was implemented using \texttt{scikit-learn} \cite{pedregosa2011scikit}, which is distributed under the BSD License. Neural network models were implemented in \texttt{PyTorch} \cite{paszke2019pytorch}, which is distributed under a BSD-style license. The differentiable ODE solves were performed using \texttt{torchdiffeq} \cite{chen2018torchdiffeq}, which is distributed under the MIT License. We used these packages through their public APIs and respected their license terms.

\section{Results}
\begin{table*}[t]
\centering
\caption{NMAE results for all model forecasts from day 3.}
\label{tab:day3_results_and_ci}
\small
\resizebox{\textwidth}{!}{%
\begin{tabular}{lcccccccccc}
\toprule
Model & Ammonia & Calcium & Glutamate & Glutamine & Lactate & Osmolality & Potassium & Sodium & VCD & Avg. Rank \\
\midrule
CART & 0.835 & 3.584 & 0.523 & 0.276 & 1.910 & 1.208 & 3.420 & 1.947 & 1.041 & 7.22 \\
Global Latent ODE & 0.300 & 0.585 & 0.448 & 0.321 & 0.408 & 0.399 & 0.633 & \textbf{0.443} & 0.468 & 4.44 \\
LKF & 1.506 & 3.347 & 1.710 & 1.805 & 1.618 & 0.658 & 1.208 & 1.089 & 0.354 & 7.00 \\
MP-JIT-FT Latent ODE & 0.273 & 0.623 & 0.325 & 0.406 & 0.439 & 0.371 & 0.548 & 0.517 & 0.436 & 4.56 \\
MP-JIT-FT GB-Latent ODE (1) & \textcolor{red}{0.228} & 0.508 & 0.413 & 0.390 & \textbf{0.327} & 0.400 & \textbf{0.414} & 0.533 & \textcolor{red}{0.291} & \textbf{3.22} \\
(1) + Controlled Variable & \textbf{0.272} & 0.600 & 0.480 & 0.445 & \textcolor{red}{0.292} & 0.469 & 0.582 & 0.498 & 0.337 & 4.22 \\
(1) + Raman & 0.312 & \textcolor{red}{0.294} & \textcolor{red}{0.214} & \textcolor{red}{0.145} & 0.380 & \textcolor{red}{0.252} & \textcolor{red}{0.355} & \textcolor{red}{0.306} & \textbf{0.322} & \textcolor{red}{1.78} \\
(1) + Controlled Variable + Raman & 0.318 & \textbf{0.437} & \textbf{0.274} & \textbf{0.153} & 0.430 & \textbf{0.320} & 0.598 & 0.466 & 0.367 & 3.56 \\
\bottomrule
\end{tabular}%
}
\end{table*}

\begin{table*}[t]
\centering
\caption{All model day 3 forecast 95\% confidence interval widths.}
\label{tab:day3_ci}
\small
\resizebox{\textwidth}{!}{%
\begin{tabular}{lccccccccc}
\toprule
Model & Ammonia & Calcium & Glutamate & Glutamine & Lactate & Osmolality & Potassium & Sodium & VCD \\
\midrule
CART & 0.131 & 0.782 & 0.143 & \textbf{0.065} & 0.332 & 0.235 & 0.692 & 0.452 & 0.491 \\
Global Latent ODE & 0.053 & 0.130 & 0.095 & 0.076 & \textcolor{red}{0.094} & 0.092 & 0.130 & \textcolor{red}{0.107} & 0.195 \\
LKF & 0.215 & 1.150 & 0.663 & 0.783 & 0.257 & 0.071 & 0.317 & 0.424 & \textcolor{red}{0.058} \\
MP-JIT-FT Latent ODE & \textbf{0.050} &    0.112 &    0.091 &    0.149 & 0.260 &    \textbf{0.065} &    0.125 &    0.137 & 0.139 \\
MP-JIT-FT GB-Latent ODE (1) & \textcolor{red}{0.040} & 0.104 & 0.122 & 0.178 & \textbf{0.147} & 0.129 & \textbf{0.093} & 0.189 & 0.248 \\
(1) + Controlled Variable & 0.069 & 0.101 & 0.350 & 0.174 & 0.179 & 0.135 & 0.133 & 0.171 & 0.080 \\
(1) + Raman & 0.058 & \textcolor{red}{0.081} & \textcolor{red}{0.052} & \textcolor{red}{0.038} & 0.272 & \textcolor{red}{0.044} & \textcolor{red}{0.061} & \textbf{0.125} & \textbf{0.063} \\
(1) + Controlled Variable + Raman & 0.055 & \textbf{0.088} & \textbf{0.054} & \textcolor{red}{0.038} & 0.400 & 0.085 & 0.131 & 0.358 & 0.070 \\
\bottomrule
\end{tabular}%
}
\end{table*}

Table~\ref{tab:day3_results_and_ci} reports the NMAE values across all target variables, with red text and bold text indicating the best and second-best performing models, respectively, for each target variable. The corresponding 95\% confidence interval widths for each model are shown in Table~\ref{tab:day3_ci}. Overall, directly applying MP-JIT-FT to the Latent ODE did not improve its average rank across targets. This may reflect overfitting from the added ratio features in the small-data setting. Adding the gated bottleneck architecture addressed the issue and achieved the best performance for Ammonia and VCD. At the variant level, MP-JIT-FT GB-Latent ODE with Raman achieves the highest average rank (i.e. the best performance on average) and outperforms the Global Latent ODE baseline on 8 out of 9 variables, with Ammonia as the only exception.


\begin{figure}[!t]
\centering
\includegraphics[width=0.8\columnwidth]{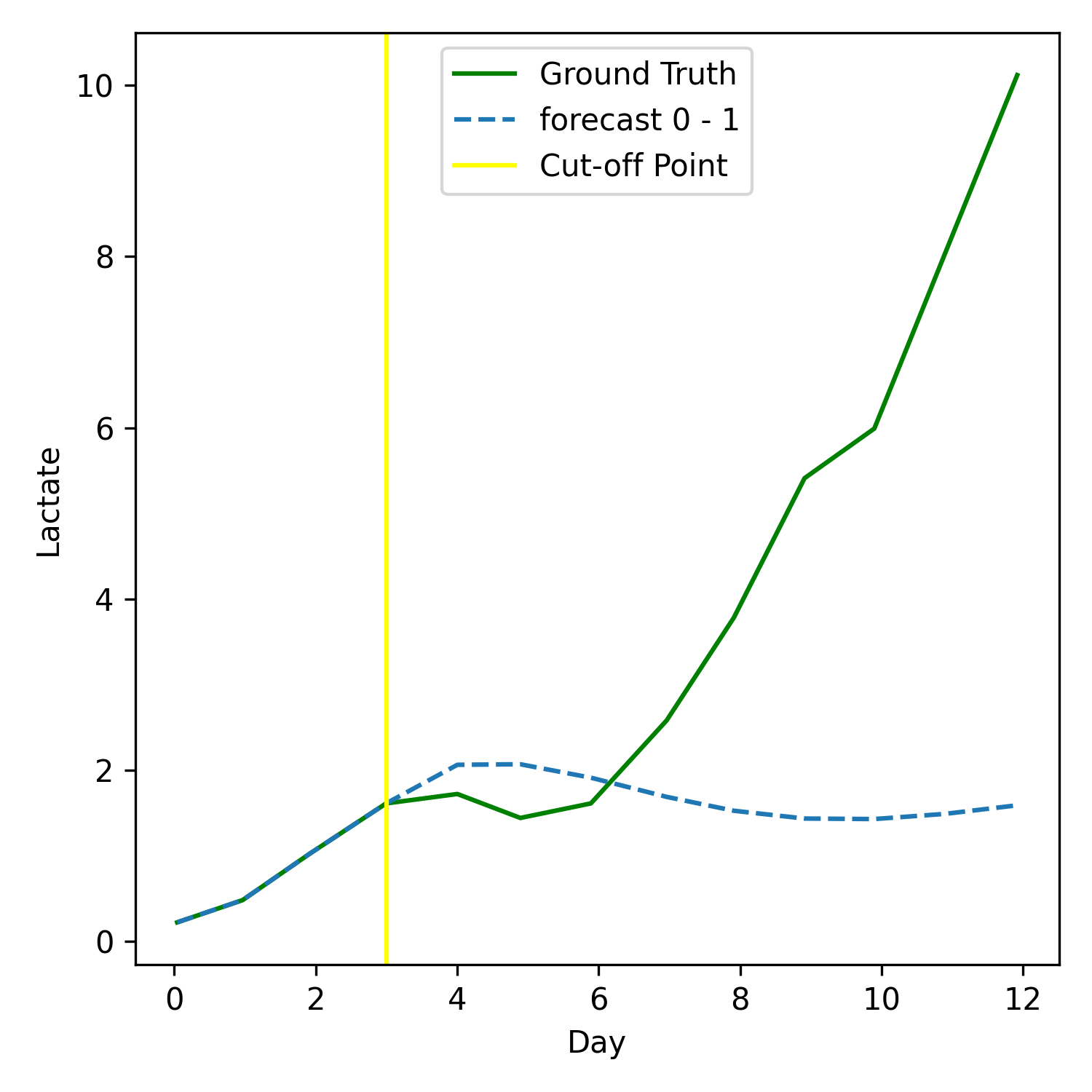}\par\medskip
\includegraphics[width=0.8\columnwidth]{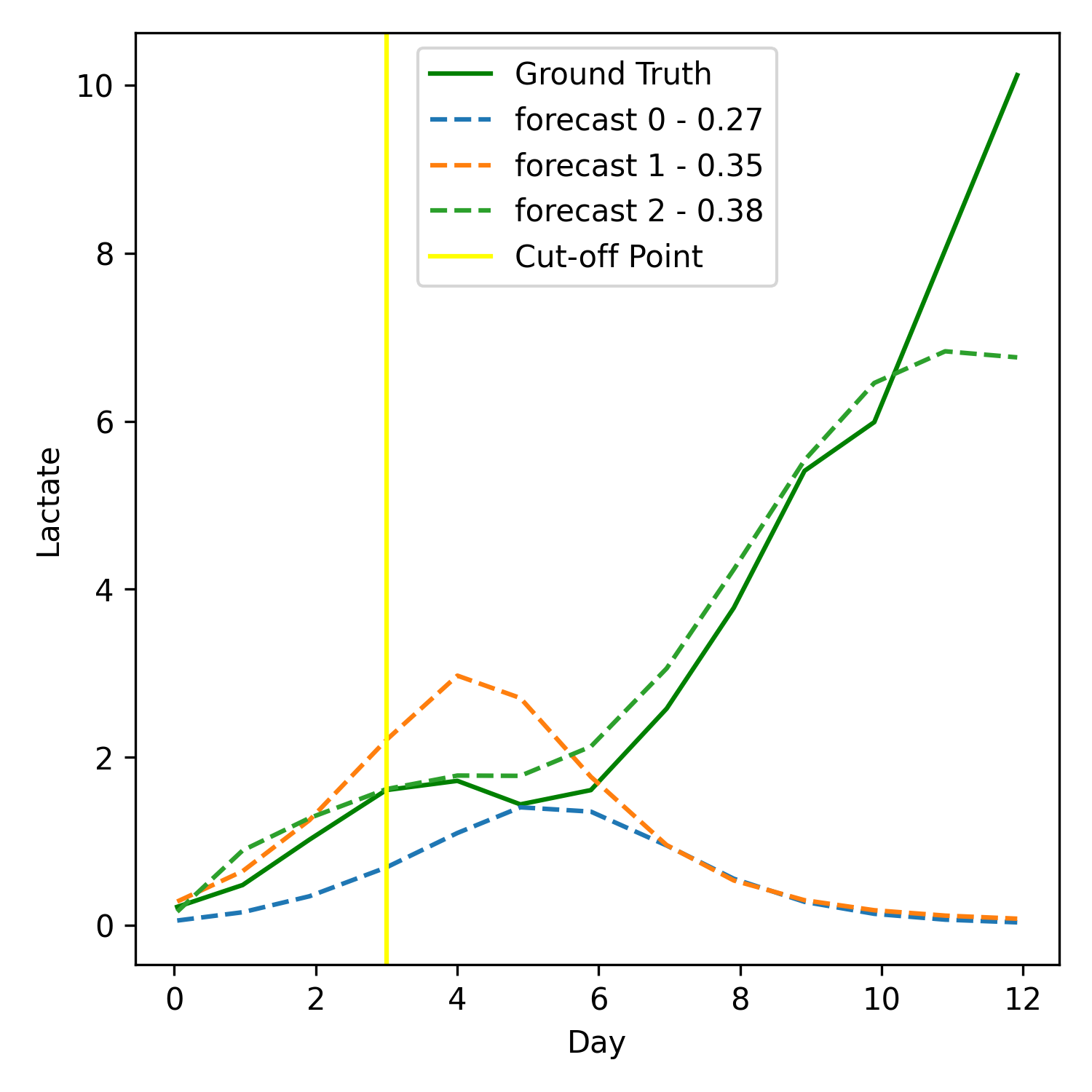}
\caption{Global Latent ODE (top) and MP-JIT-FT GB-Latent ODE (bottom).}
\label{fig:example_lactate_mpjitft}
\end{figure}

\begin{figure}[!t]
\centering
\includegraphics[width=0.8\columnwidth]{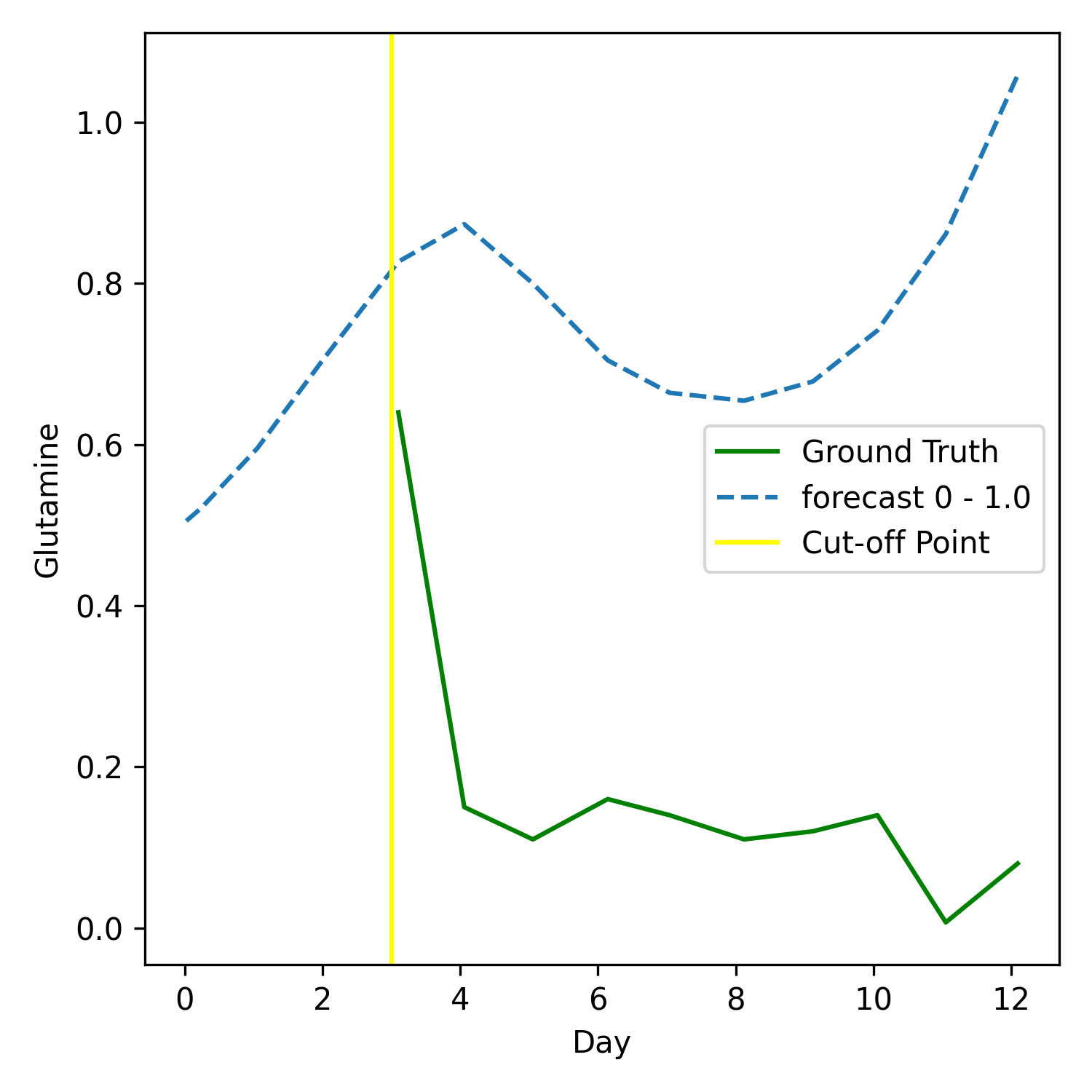}\par\medskip
\includegraphics[width=0.8\columnwidth]{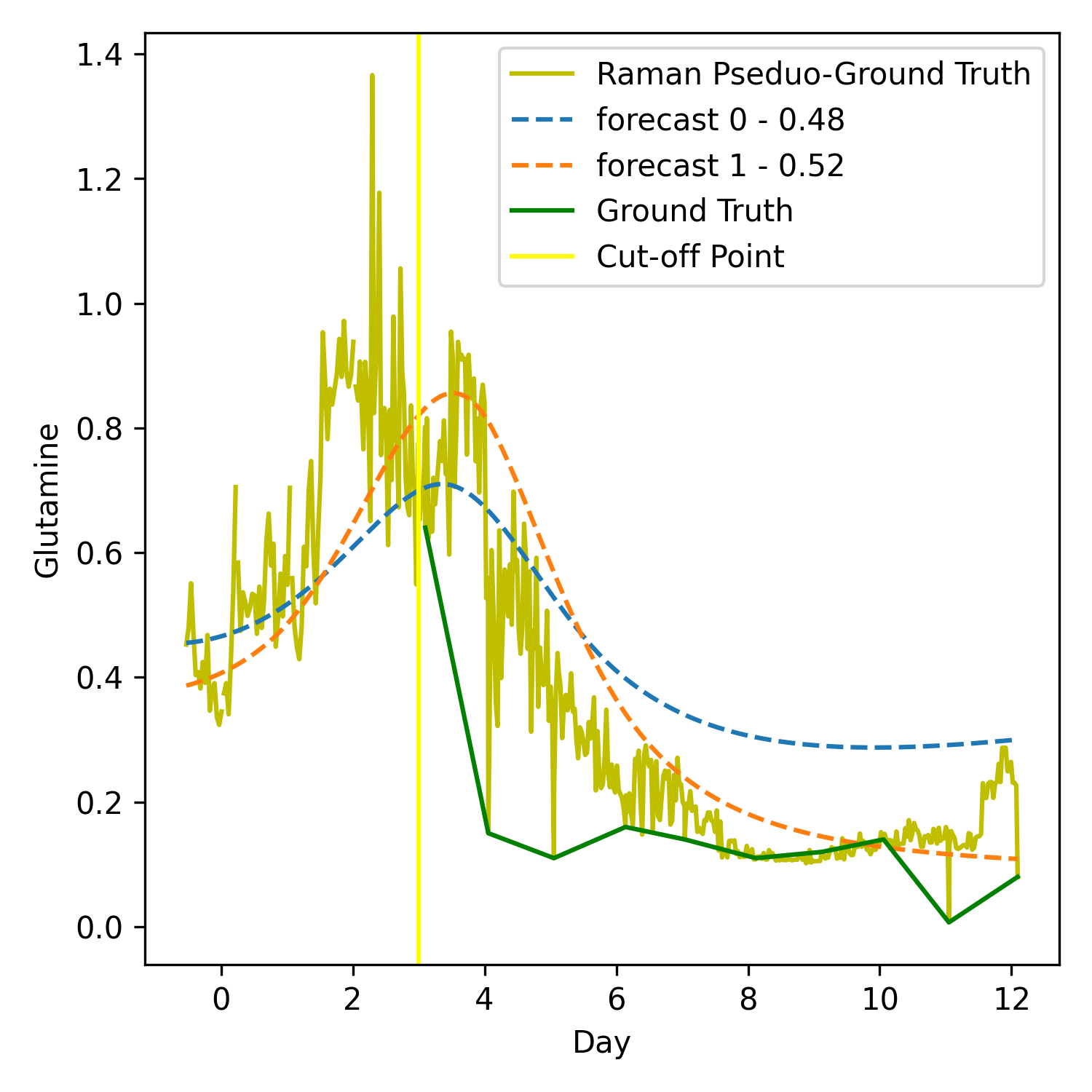}
\caption{GB-Latent ODE without (top) and with (bottom) Raman data fusion.}
\label{fig:no_prefix_raman}
\vspace{-10pt}
\end{figure}

In contrast, adding controlled variables mainly improves Lactate, while most other targets degrade relative to MP-JIT-FT GB-Latent ODE without controlled variables. This is biologically plausible, as the controlled variables considered here mainly describe gas transfer, acid-base conditions, and operating state, which are closely linked to lactate metabolism in CHO cultures \cite{brunner2018elevated,anane2021scale,gagnon2011high,park2022energy}. The degradation on other targets does not imply biological irrelevance, as pH, pO$_2$, and pCO$_2$ also affect CHO physiology, growth, and amino-acid metabolism \cite{brunner2017investigation}. Instead, these results suggest that controlled variables have target-dependent predictive value and may be harder to exploit in small heterogeneous datasets due to increased dimensionality and collinearity. Next, we analyse the relationship between target-dependent gains from the proposed methods and local trajectory divergence.



\subsection{Local Divergence Slope \& Local Divergence Offset}
MP-JIT-FT is designed for settings where trajectories with similar prefixes may diverge in the future. Therefore, its improvement over the global Latent ODE should increase with local post-cutoff divergence. To test this, we group trajectories with similar pre-cutoff behaviour and measure two forms of future separation: the \emph{Local Divergence Offset} (LDO), which captures the fitted post-cutoff separation level after accounting for separation at the cutoff, and the \emph{Local Divergence Slope} (LDS), which measures how quickly this separation grows. This construction is inspired by Lyapunov-style separation analysis \cite{rosenstein1993}. The technical computation details about LDO \& LDS are discussed below:

Let $c$ denote the cutoff time. For each trajectory, we split the observed series into a prefix segment up to $c$ and a future segment from $c$ onward, with the value at the cutoff obtained by interpolation when necessary. To group trajectories with similar prefix behaviour, we compute pairwise distances between prefix segments using root-mean-square error over their common overlap interval. The pairwise distances are then supplied to a $k$-medoids procedure, which partitions the trajectories into clusters of similar pre-cutoff behaviour.

For a given cluster $\mathcal{C}$, we then consider all unordered within-cluster trajectory pairs $(i,j)$ with $i,j \in \mathcal{C}$ and $i \neq j$. Let $y_i(c)$ and $y_j(c)$ denote the values of trajectories $i$ and $j$ at the cutoff. Their initial separation is defined as
\begin{equation}
d_{ij}(0) = \left| y_i(c) - y_j(c) \right|.
\end{equation}

Next, both trajectories are evaluated on their common future time interval after the cutoff. At each future lag $\tau > 0$, the pairwise separation is
\begin{equation}
d_{ij}(\tau) = \left| y_i(c+\tau) - y_j(c+\tau) \right|.
\end{equation}
Using these quantities, we define the log-separation growth for pair $(i,j)$ at lag $\tau$ as
\begin{equation}
g_{ij}(\tau)
=
\log\left(
\frac{d_{ij}(\tau)+\varepsilon}{d_{ij}(0)+\varepsilon}
\right),
\end{equation}
where $\varepsilon > 0$ is a small constant introduced for numerical stability. For each future lag $\tau$, we average this quantity across all within-cluster trajectory pairs with valid overlap at that lag:
\begin{equation}
\bar{g}_{\mathcal{C}}(\tau)
=
\frac{1}{|\mathcal{P}_{\mathcal{C}}(\tau)|}
\sum_{(i,j)\in\mathcal{P}_{\mathcal{C}}(\tau)}
g_{ij}(\tau),
\end{equation}
where $\mathcal{P}_{\mathcal{C}}(\tau)$ denotes the set of valid within-cluster pairs at lag $\tau$.

Finally, a linear model is fitted to the cluster-level mean log-growth curve using ordinary least squares over all available future lags as shown in the equation below:
\begin{equation}
\bar{g}_{\mathcal{C}}(\tau) \approx a_{\mathcal{C}} \tau + b_{\mathcal{C}},
\end{equation}
The fitted slope \(a_{\mathcal C}\) and intercept \(b_{\mathcal C}\) are defined
as the Local Divergence Slope (LDS) and Local Divergence Offset (LDO) of
cluster \(\mathcal C\), respectively.LDS represents how fast similar local trajectories diverge in the future, while LDO represents the fitted post-cutoff log-separation offset. A positive LDS and LDO indicate that trajectories with similar prefix behaviour tend to separate after the cutoff, while values near zero suggest weak or no divergence, and negative values suggest that trajectories converge after the cutoff.

\subsection{Improvement From MP-JIT-FT}
\begin{figure*}[!t]
\centering
\subfloat[MP-JIT-FT: LDO]{\includegraphics[width=0.45\textwidth]{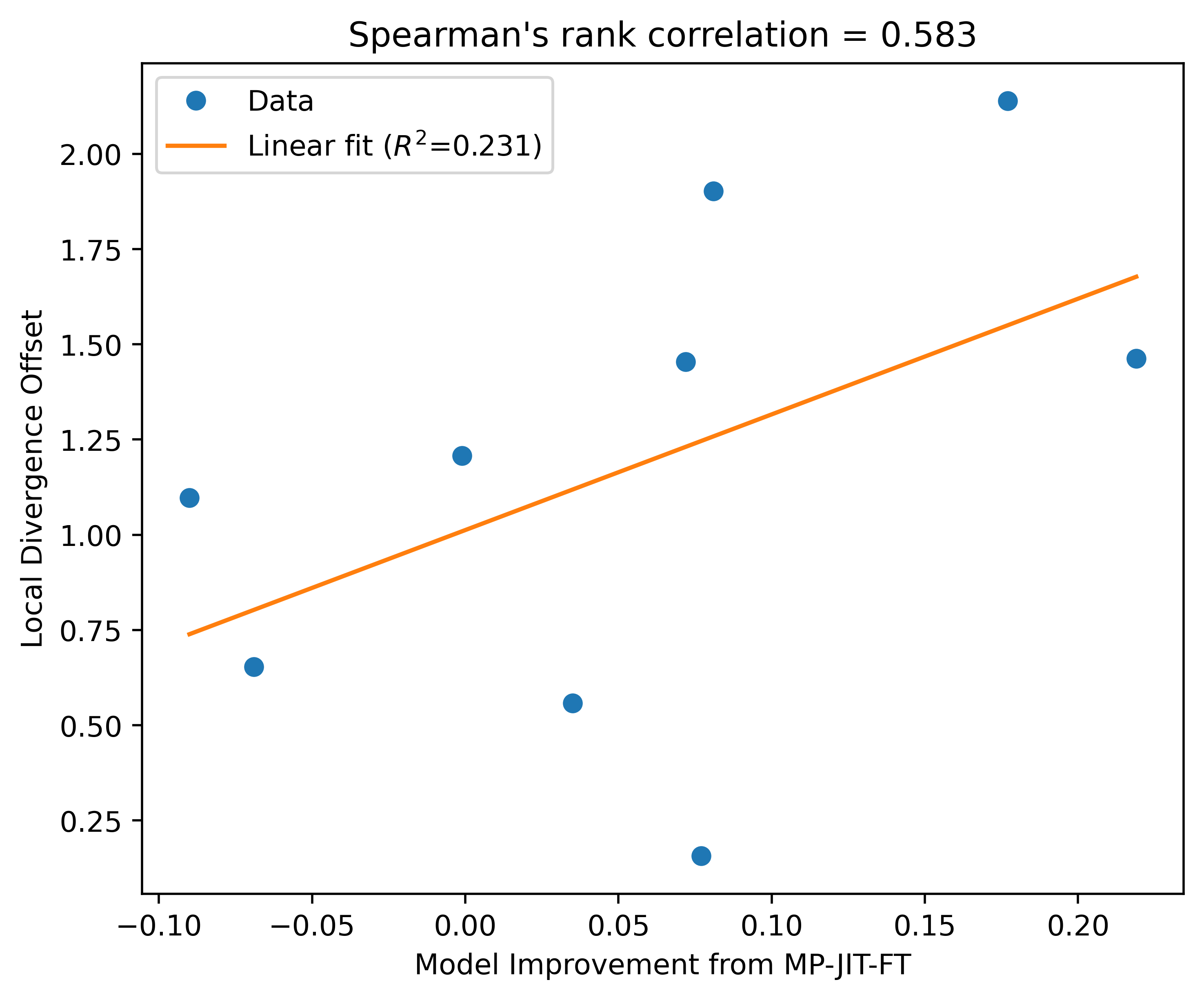}%
\label{fig:ald_mpjit}}
\hfil
\subfloat[MP-JIT-FT: LDS]{\includegraphics[width=0.45\textwidth]{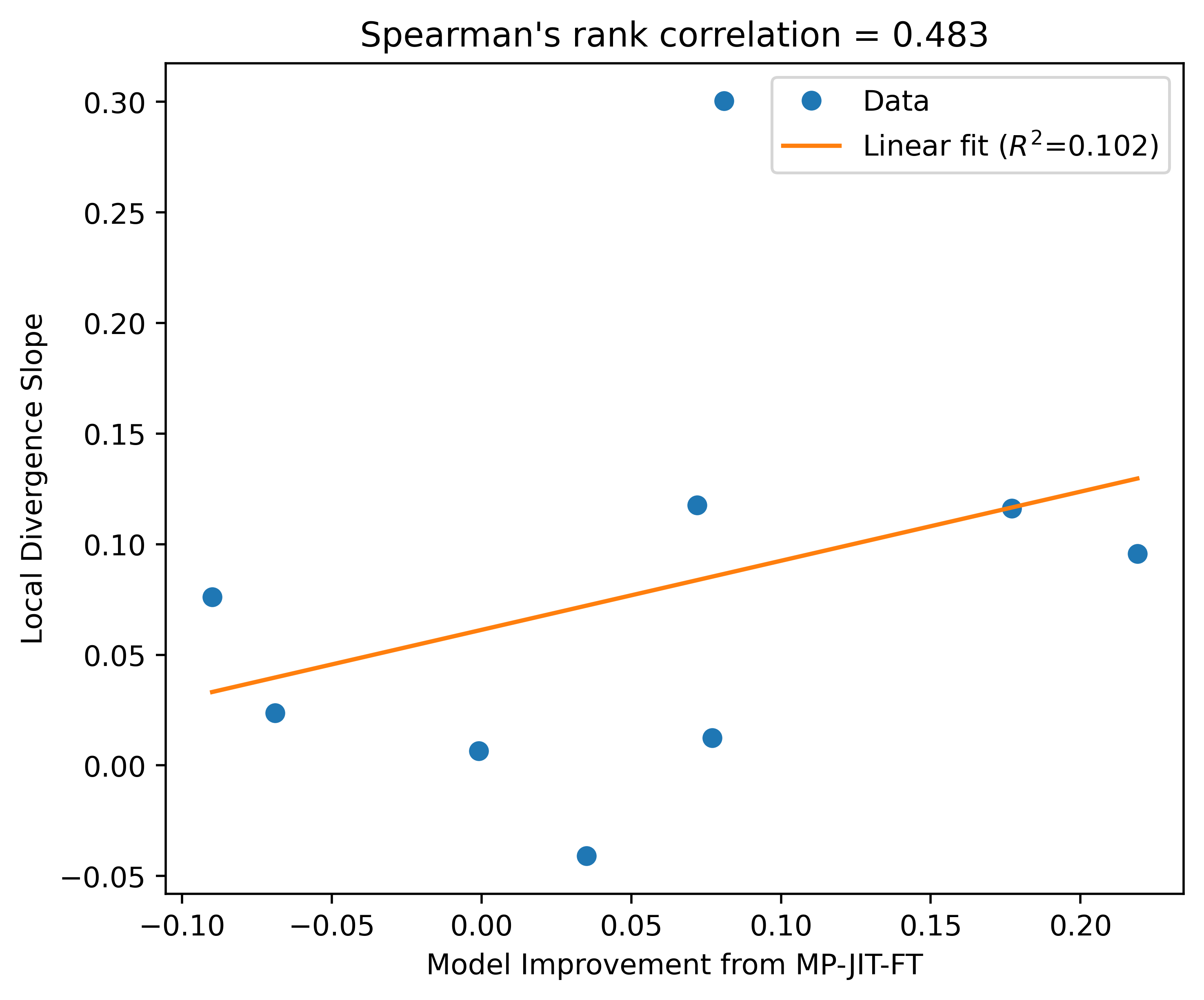}%
\label{fig:lds_mpjit}}
\vspace{0pt}
\subfloat[Raman Data Fusion: LDO]{\includegraphics[width=0.45\textwidth]{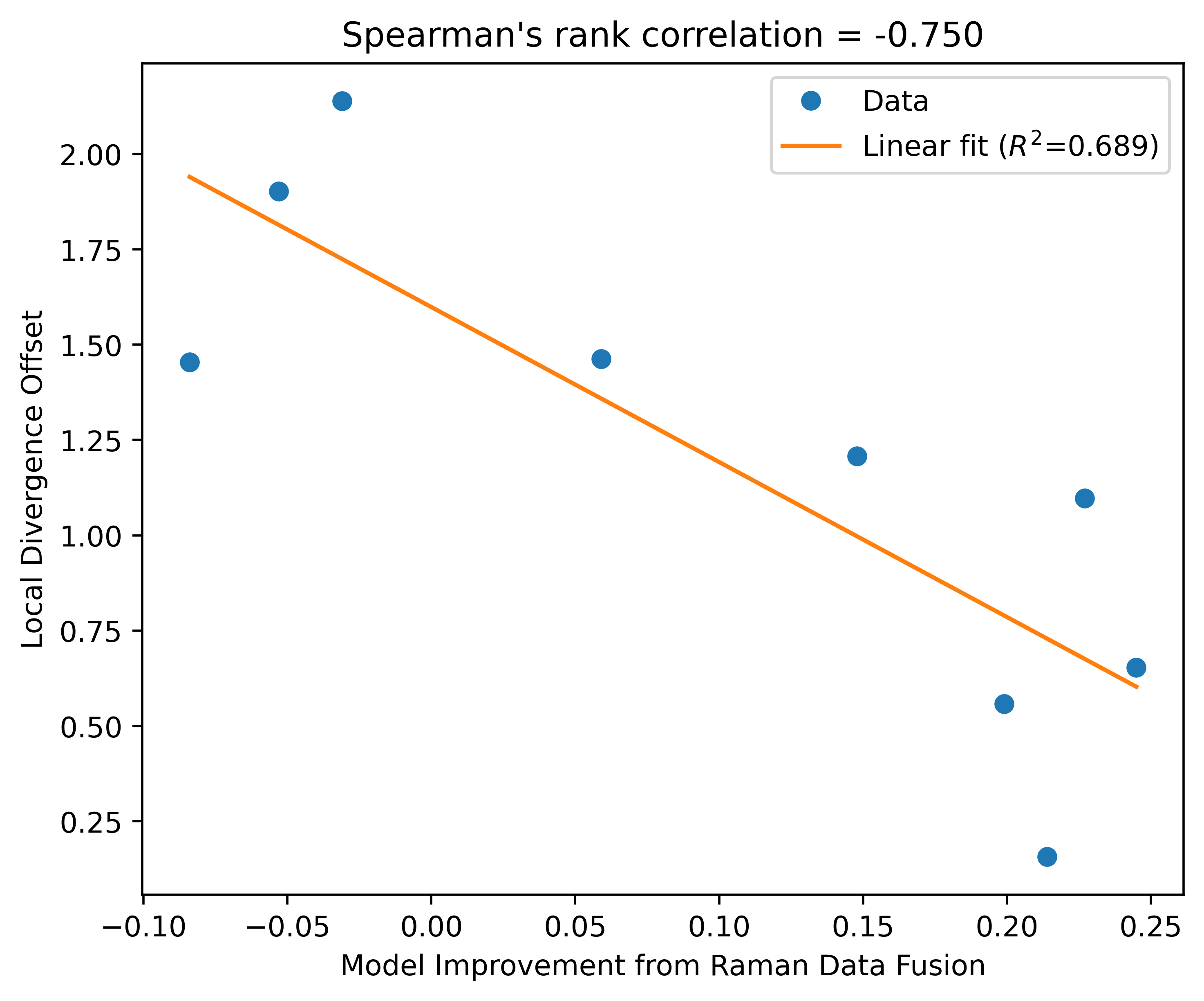}%
\label{fig:ald_raman}}
\hfil
\subfloat[Raman Data Fusion: LDS]{\includegraphics[width=0.45\textwidth]{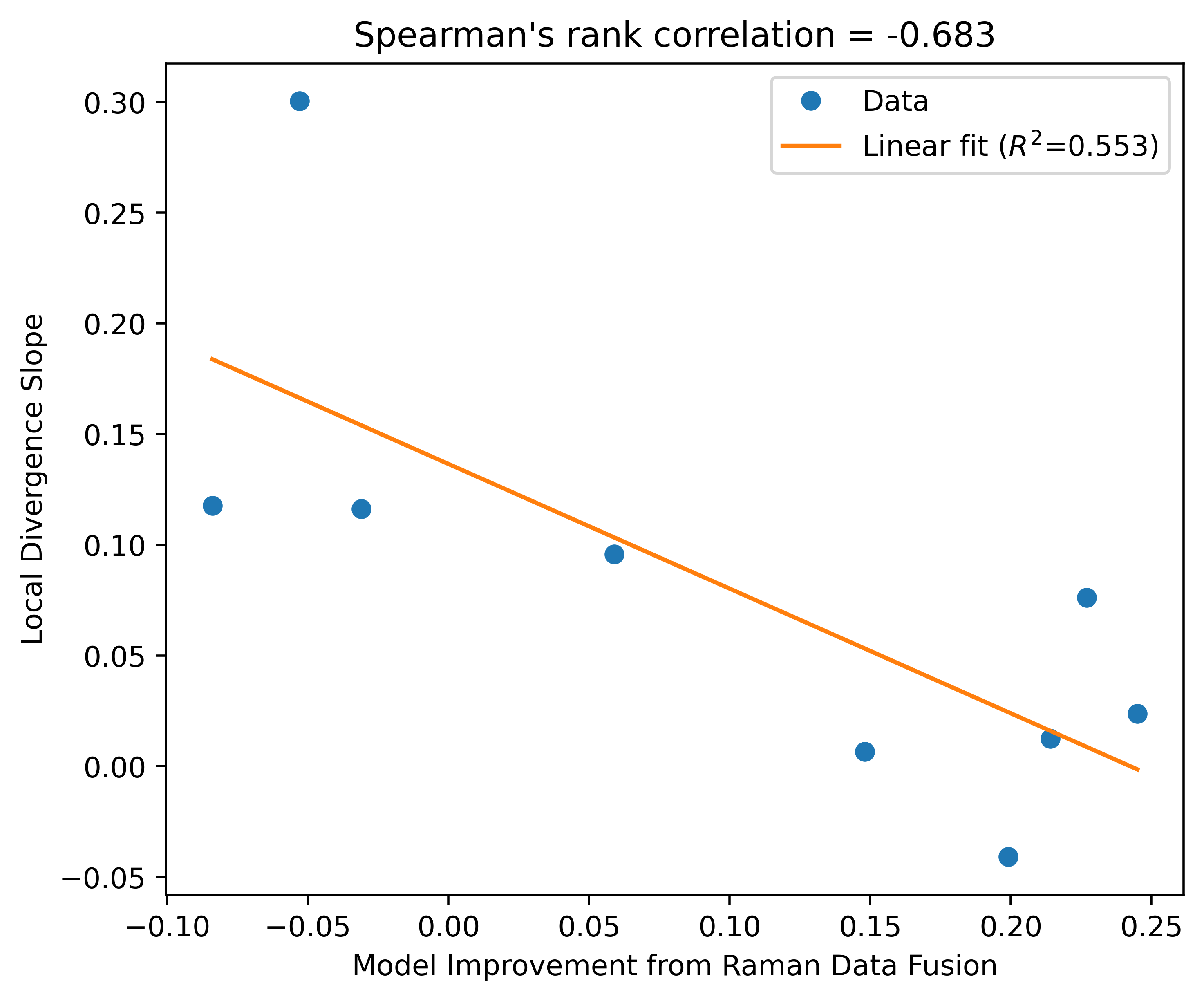}%
\label{fig:lds_raman}}
\caption{Spearman's rank correlation and $R^2$ between model improvement and local divergence metrics. Top row: MP-JIT-FT improvement. Bottom row: Data fusion improvement. Left column: LDO. Right column: LDS. Seed = 33.}
\label{fig:ald_lds_all}
\end{figure*}

For each target variable, we computed cluster-size-weighted average LDO and LDS and correlated them with the NMAE improvement of MP-JIT-FT GB-Latent ODE over the global Latent ODE. Across 100 clustering initialisations, the average Spearman correlations with LDS and LDO were \(0.423\) and \(0.545\), respectively, with standard deviations of \(0.091\) and \(0.022\), and average p-values of \(0.270\) and \(0.130\). While the p-value is limited by the small sample size, positive correlations suggest that MP-JIT-FT provides more gains when locally similar prefixes diverge, such as the scenario shown in Fig.~\ref{fig:example_lactate_mpjitft} (path confidences shown in the legend). The corresponding correlation plots are shown in Fig.~\ref{fig:ald_lds_all}.

\subsection{Improvement From Raman Data Fusion}
We performed the same analysis for the improvement from Raman data fusion. Across 100 clustering initialisations, the average Spearman correlations between Raman-related improvement and LDS and LDO were \(-0.768\) and \(-0.739\), respectively, with standard deviations of \(0.042\) and \(0.029\), and average p-values of \(0.017\) and \(0.024\). Despite a small sample size, the p-value remains low, suggesting Raman data fusion is less beneficial when future divergence is high, and vice versa. Thus, Raman pseudo-observations are most useful when early-run dynamics are locally representative of later behaviour. Because they are generated by a separate soft-sensor model and given the same loss weight as offline assays, they can reinforce dense early-prefix patterns. This helps in locally homogeneous regimes but can encourage overfitting when similar prefixes diverge.

Raman pseudo-observations are also beneficial when offline measurements are unavailable or limited before day 3. In this setting, forecasting without prior measurements is challenging, as shown at the top of Fig.~\ref{fig:no_prefix_raman}. While incorporating Raman pseudo-observations provided the model with some guidance, which substantially improved the forecasts, as shown at the bottom of the figure.

\section{Limitations}
Beyond the limited dataset size and benchmark coverage, this study has three main limitations.
\paragraph{Weights on Raman Pseudo-observations}
Raman pseudo-observations are assigned the same loss weight as offline measurements by setting \(\lambda=1\). This avoids additional tuning but assumes that dense Raman-based predictions and sparse offline assays have equal reliability, despite different noise levels. Our results suggest this fixed weighting is not always optimal: Raman data fusion helps when trajectories are locally homogeneous, but can hurt when similar early prefixes diverge after the forecast origin. Future work should explore uncertainty-aware or adaptive weighting, for example, by estimating \(\lambda\) from Raman validation error, prediction intervals, target-specific reliability, or local divergence metrics such as LDS and LDO.
\paragraph{JITL Retrieval and Clustering Hyperparameters}
MP-JIT-FT currently uses manually specified hyperparameters, including the number of retrieved neighbours \(k\), the clustering percentile threshold, and target-specific retrieval weights. These choices make the method simple and reproducible, but may not be optimal across variables, operating regimes, or datasets with different levels of local heterogeneity. Future work could replace these fixed choices with data-driven alternatives such as meta-learning, or ensemble-based approaches \cite{PENG2025157}.
\paragraph{Training and Inference Time}
As shown in Table~\ref{table:runtime}, MP-JIT-FT has substantially higher inference cost than global baselines such as CART, LKF, and Latent ODE because each target requires target-specific neighbour retrieval, clustering, and cluster-specific fine-tuning. As a result, although each fine-tuned model can output multiple variables, only the selected target prediction is retained, and the procedure must be repeated for each target variable. Raman data fusion further increases runtime by adding dense pseudo-observations. On the other hand, the wall-clock time is substantially lower through parallel processing, since forecasts for different targets and candidate paths are largely independent. Thus, this runtime is acceptable for our application, where forecasts do not need to be refreshed frequently, but it may limit use in time-critical applications requiring frequent inference.

\begin{table*}[!t]
\centering
\caption{Average training and inference runtime under 38-run cross-validation, reported in seconds. Experiments used a 2025 Mac Studio with M4 Max chip, 14-core CPU, 32-core GPU, 36GB RAM.}
\label{table:runtime}
\small
\resizebox{0.85\textwidth}{!}{%
\begin{tabular}{lrrr}
\toprule
Model
& Avg. Train
& Avg. Inference
& Avg. Wall-clock Time \\
\midrule
CART
& 0.026
& 0.003
& 0.275 \\

LKF
& 37.076
& 0.004
& 3.287 \\

Latent ODE
& 970.953
& 0.008
& 84.211 \\

MP-JIT-FT Latent ODE
& 718.491
& 997.479
& 134.714 \\

MP-JIT-FT GB-Latent ODE (1)
& 803.569
& 1043.253
& 145.004 \\

(1) + Controlled Variables
& 841.456
& 1086.680
& 152.725 \\

(1) + Raman
& 17656.359
& 14149.444
& 2557.073 \\

(1) + Raman \& Controlled Variables
& 18520.096
& 14677.583
& 2668.258 \\
\bottomrule
\end{tabular}%
}
\end{table*}

\section{Conclusion}
We presented an adaptive framework for forecasting irregular and heterogeneous mammalian cell-culture trajectories that combines three components: a Gated Bottleneck Latent ODE (GB-Latent ODE) that compresses sparse, high-dimensional, partially observed inputs through learnable variable-wise gating and a mask-aware bottleneck; a Multi-Path Just-In-Time Fine-Tuning (MP-JIT-FT) framework that retrieves locally similar runs, clusters them into candidate future regimes, and fine-tunes a separate model per regime to emit multiple confidence-scored forecast paths instead of a single averaged prediction; and an optional Raman data fusion mechanism that converts dense Raman spectra into pseudo-observations to enrich the sparse offline assays.

Across 38 fed-batch 5L bioreactor runs spanning 14 experimental conditions, the framework improved forecasting accuracy for most target variables, and MP-JIT-FT GB-Latent ODE with Raman fusion achieved the best average rank, outperforming the global Latent ODE baseline on 8 of 9 variables. A local-divergence analysis showed that the multi-path gains are largest when locally similar prefixes diverge after the forecast origin, while Raman fusion is most useful when early dynamics are representative of later behaviour. The improvements are therefore target-dependent rather than uniform: for instance, adding controlled variables helped lactate but degraded several other targets, reflecting the difficulty of exploiting higher-dimensional inputs in small, heterogeneous datasets.

These findings point to clear next steps: replacing the fixed Raman pseudo-observation weight with uncertainty- or divergence-aware weighting, and replacing the manually specified retrieval and clustering hyperparameters with data-driven alternatives such as meta-learning or ensemble-based approaches. Finally, while the framework is motivated by a specific bioprocessing application, it is designed to address the more general problem of multiple-future forecasting under partial observability. We intend to test it on diverse multi-modal trajectory-forecasting problems beyond cell-culture processes and, by releasing it as fully open source, encourage and welcome others to apply and extend it in such diverse domains and scenarios.

\section*{CRediT authorship contribution statement}
\textbf{Johnny Peng:} Conceptualisation, Methodology, Investigation, Software, Validation, Visualisation, Writing – original draft. \textbf{Thanh Tung Khuat:} Conceptualisation, Methodology, Investigation, Validation, Writing – review \& editing. \textbf{Ellen Otte:} Conceptualisation, Investigation, Validation, Supervision, Writing – review. \textbf{Katarzyna Musial:} Conceptualisation, Investigation, Validation, Supervision, Writing – review \& editing. \textbf{Bogdan Gabrys:} Conceptualisation, Methodology, Investigation, Validation, Project administration, Funding acquisition, Writing – review \& editing. 

\section*{Declaration of Competing Interest}
Ellen Otte is an employee of CSL Innovation Pty Ltd. The other authors declare no competing interests, including no known competing financial interests or personal relationships that could have appeared to influence the work reported in this paper.

\section*{Acknowledgements}
This research was supported under the Australian Research Council's Industrial Transformation Research Program (ITRP) funding scheme (project number IH210100051). The ARC Digital Bioprocess Development Hub is a collaboration between The University of Melbourne, University of Technology Sydney, RMIT University, CSL Innovation Pty Ltd, Cytiva (Global Life Science Solutions Australia Pty Ltd) and Patheon Biologics Australia Pty Ltd.

\bibliographystyle{IEEEtran}
\bibliography{neurips}

\end{document}